\documentclass{article}

 \usepackage[preprint]{neurips_2026}

\usepackage[utf8]{inputenc} % allow utf-8 input
\usepackage[T1]{fontenc}    % use 8-bit T1 fonts
\usepackage{microtype}
\usepackage{graphicx}
\usepackage{subcaption}
\usepackage{booktabs} % for professional tables
\usepackage{floatrow}
\usepackage{wrapfig}
\usepackage{hyperref}
\usepackage{url}
\usepackage{subcaption}
\usepackage{amsmath}
\usepackage{bbm}
\usepackage[utf8]{inputenc} % allow utf-8 input
\usepackage[T1]{fontenc}    % use 8-bit T1 fonts
\usepackage{hyperref}       % hyperlinks
\usepackage{url}            % simple URL typesetting
\usepackage{booktabs}       % professional-quality tables
\usepackage{amsfonts}       % blackboard math symbols
\usepackage{nicefrac}       % compact symbols for 1/2, etc.
\usepackage{microtype}      % microtypography
\usepackage{xcolor}         % colors
\usepackage{graphicx}
\usepackage{enumitem}
\usepackage{xcolor}
\usepackage{blindtext}
\usepackage{comment}
\usepackage{floatrow}
\usepackage{mathtools}

\usepackage{minted}
\usepackage{algorithm}
\usepackage{pifont}
\newcommand{\xmark}{\ding{55}}%
\usepackage{tabularx}
\usepackage{longtable}
\usepackage{multirow}
\usepackage{verbatim} 
\usepackage{afterpage}
\usepackage{subcaption}
\usepackage{floatrow}
\newfloatcommand{capbtabbox}{table}[][\FBwidth]
\title{AdaDim: Dimensionality Adaptation for SSL Representational Dynamics}

\author{%
  Kiran Kokilepersaud \\
  Department of Electrical and Computer Engineering\\
  Georgia Institute of Technology\\
  Atlanta, GA 30363 \\
  \texttt{kkokilepersaud@gmail.com} \\
  \And
  Mohit Prabhushankar \\
  Georgia Institute of Technology\\
  Atlanta, GA 30363 \\
  \texttt{mohit.p@gatech.edu} \\
  \And
  Ghassan AlRegib \\
  Georgia Institute of Technology\\
  Atlanta, GA 30363 \\
  \texttt{alregib@gatech.edu} \\
}

\begin{document}

\maketitle

\begin{abstract}
  A key factor in effective Self-Supervised Learning (SSL) is preventing dimensional collapse, where higher-dimensional representation spaces span a lower-dimensional subspace. Therefore, SSL optimization strategies involve guiding a model to produce a representation space with a higher dimensionality through objectives that encourage feature decorrelation or sample uniformity. A higher dimensionality indicates that the representation space has greater feature diversity which is useful for generalization to downstream tasks. In addition to dimensionality optimization, SSL algorithms also utilize a projection head that maps the representation space to an embedding space where the optimization objective is applied. Recent work has characterized the projection head as a filter of noisy or irrelevant features by reducing the mutual information between the representation and embedding spaces.  Therefore, the current literature views a good SSL representation space as one with a high dimensionality and a low mutual information between each respective space.  However, we directly oppose this perspective and instead assert that the best performing SSL model arrives at an ideal balance between both dimensionality and mutual information. This conclusion is supported through a novel training dynamics analysis that reveals that increases in dimensionality due to feature decorrelation at the start of training lead to correspondingly higher mutual information, while increases in dimensionality due to samples distributing uniformly in a high-dimensional space at the end of training cause mutual information to plateau or decrease. To leverage these dynamics, we introduce AdaDim, a method that  adaptively balances between dimensionality and mutual information by identifying situations where both families of SSL approaches are better utilized. We show that AdaDim results in performance exceeding common SSL baselines by as much as 3\% accuracy on common SSL benchmark tasks.
\end{abstract}

\section{Introduction}
\label{sec:intro}

Self-supervised Learning (SSL) \cite{uelwer2025survey} algorithms approach or surpass fully supervised strategies on a wide variety of benchmark tasks \cite{chen2020improved,chen2020simple,dwibedi2021little,zbontar2021barlow,bardes2021vicreg,chen2021exploring}. SSL optimization generally involves an invariance loss that ensures representations of similar samples align with each other and a mechanism to prevent dimensional collapse \cite{jing2021understanding}. Dimensional collapse refers to the phenomena where high dimensional representations span a lower-dimensional subspace. Therefore, to prevent dimensional collapse, a wide variety of works \cite{garrido2023rankme,agrawal2022alpha,thilak2023lidar} suggest that good SSL representations ($R$) should have a higher overall dimensionality. 
We follow the convention of previous works \cite{garrido2023rankme} that analytically measures the dimensionality of the representation space $H(R)$ through the effective rank metric \cite{roy2007effective}. Effective rank quantifies the entropy of singular values of $R$ and provides a matrix approximation of entropy ~\cite{von2018mathematical,renyi1961measures}. For this reason, we refer to dimensionality as $H(R)$ since eigenvalues approaching a uniform distribution reflect the spread of samples along higher dimensional feature directions.(Additional discussion and details of all metrics can be found in Section~\ref{sec:metrics}.) In practice, optimizing for higher dimensionality is either done through a dimension contrastive approach \cite{garrido2022duality} that encourages feature decorrelation or through a sample-contrastive method that promotes a uniform spread of sample representations \cite{wang2020understanding}. Alongside a term to promote dimensionality, all SSL methods utilize a projection head that maps $R$ to a lower dimensional embedding space $Z$ where the SSL optimization objective is applied. Recent work \cite{ouyang2025projection} has characterized the purpose of the projection head as a filter that removes spurious features, thus lowering the mutual information $I(R;Z)$ that we measure through a matrix entropy approximator \cite{renyi1961measures,zhang2023matrix}. In general, lower $I(R;Z)$ reflects representations varying only in feature directions that correspond well with task-relevant semantic concepts, while higher $H(R)$ corresponds to a greater degree of feature diversity. Together, these works imply that a good SSL representation space should have a high $H(R)$ and  a low $I(R;Z)$. 

\begin{figure*}[h!]
    \centering
     \includegraphics[width = \linewidth]{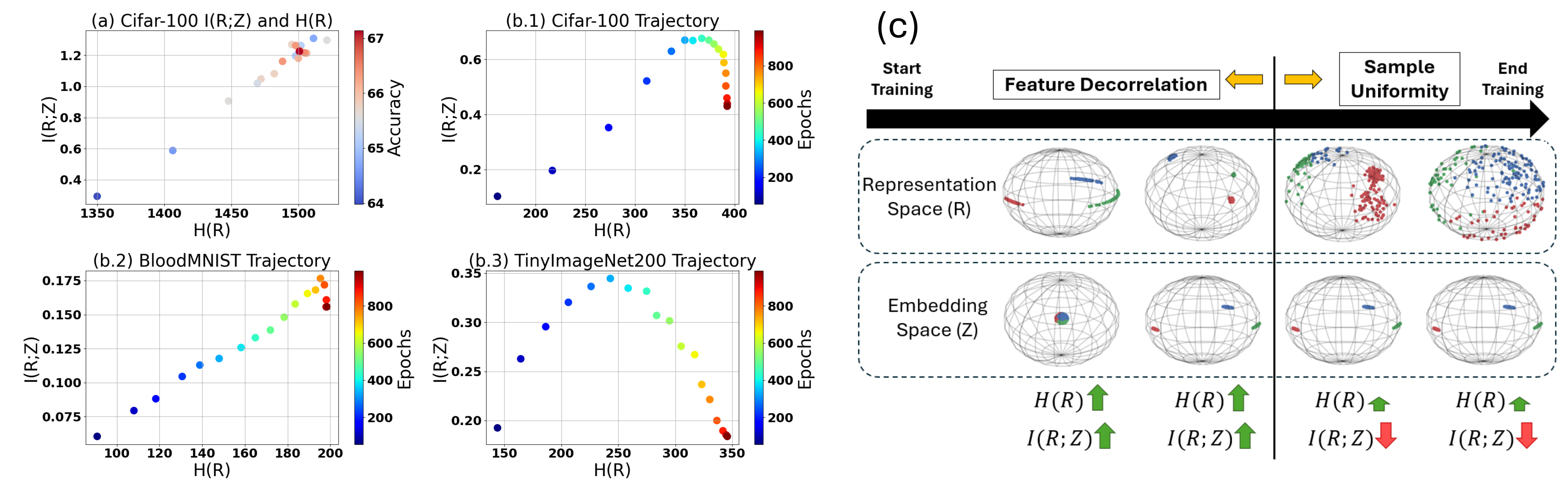}
    \caption{a) This figure shows how performance varies for 20 different pre-trained  ResNet-50 models as a function of $H(R)$ and $I(R;Z)$. b.1) - b.3) shows how $H(R)$ and $I(R;Z)$ vary across training of a ResNet-18 encoder with SimCLR \cite{chen2020simple} for 1000 epochs on three different datasets.  c) This toy graphic shows how the representation space ($R$) and embedding space ($Z$) of a toy 3D dataset changes when following SSL training dynamics. We visualize changes to $H(R)$ and $I(R;Z)$.}
    \label{fig:intro}
\end{figure*}

However, this view of SSL is lacking in terms of an understanding of the underlying training dynamics that influences the relationship between both terms. For example, in part a) of Figure \ref{fig:intro}, we show how the final $H(R)$ and $I(R;Z)$ arrived at the end of training influences downstream performance.  In this Figure, we train 20 different models with slightly different hyperparameters on a ResNet-50 \cite{he2016deep} model for 400 epochs on Cifar-100. We find that the best performing model is not the one with the highest $H(R)$ or lowest $I(R;Z)$, but instead approaches a specific $H(R)$ and $I(R;Z)$ value where downstream performance is maximized. \textbf{Thus, our first claim is that the best performing SSL representations arrive at a balance between both $H(R)$ and $I(R;Z)$ such that there is enough feature diversity for the task of interest, but not so much that $R$ contains irrelevant noise.} This claim directly opposes existing literature \cite{garrido2023rankme,agrawal2022alpha,thilak2023lidar} that only considers some associated measurement of the $H(R)$ reached at the end of training as an indicator of downstream model performance. 

We also analyze the representational dynamics that cause this behavior. In parts b.1) - b.3) of Figure \ref{fig:intro}, we show how $H(R)$ and $I(R;Z)$ evolve over the course of SimCLR \cite{chen2020simple} training on a ResNet-18 model for 1000 epochs across 3 distinct datasets. While $H(R)$ generally increases throughout training, as expected by the current literature, $I(R;Z)$ does not directly decrease and instead goes through distinct phases of increasing, plateauing, and decreasing. In part c), we show a toy example to visualize the dynamics causing this behavior. In this Figure, we have 200 samples distributed within a fictitious 3D spherical representation space. At the start of training, $H(R)$ increases by projecting $R$ onto a higher dimensional space by mapping from a 2D plane to the surface of the sphere. $Z$ correspondingly projects from a 1D to 2D space. This phase corresponds to feature decorrelation where both $R$ and $Z$ increase the number of dimensions in which they vary which causes $I(R;Z)$ to increase as both spaces are projecting to higher dimensions. However, later in training, $H(R)$ has fewer dimensions in which to project into and further increases in $H(R)$ are caused by samples distributing uniformly within the space. This change in sample spread is not reflected to the same degree in $Z$ which causes $I(R;Z)$ to decrease. Thus, our \textbf{second claim is that feature decorrelation at the start of training leads to higher $I(R;Z)$, while samples uniformly spreading across higher dimensions at the end of training causes $I(R;Z)$ to plateau or decrease.} This observation is significant in the context of SSL literature as it opposes previous analyses that  characterized both sample uniformity and feature decorrelation losses as equivalent under certain conditions \cite{garrido2022duality}. However, more recent theoretical work \cite{shwartz2023information}, suggested the development of SSL strategies that appropriately apply the divergent distributional characteristics of both approaches. Therefore, \textbf{our third claim is that both feature decorrelation and sample uniformity based losses are optimal at different points in training that depend on the representational dynamics of unique data settings}. We propose an SSL method called AdaDim that takes advantage of the discussed training dynamics to adaptively balance increasing $H(R)$ through feature decorrelation and sample uniformity as well as gradual regularization of $I(R;Z)$. The contributions of our work are as follows:

\begin{enumerate}
    \item We theoretically and empirically demonstrate that the relationship between $H(R)$ and $I(R;Z)$ can characterize SSL training dynamics as well as downstream performance through both Gaussian and information theoretic analyses. 
    \item We show that both feature decorrelation and sample uniformity based SSL losses have advantages at different points in training that require balancing between both during optimization.
    \item We develop a dimension adaptive method (AdaDim) that takes advantage of our discovered training dynamics to regularize the training process to balance both $H(R)$ and $I(R;Z)$. We demonstrate performance improvements in comparison with state of the art methods without needing expensive architectural strategies. However, we also show that our method can leverage these techniques for further performance gains.
\end{enumerate}

\section{Related Works}
\label{sec:related}

\paragraph{Sample vs. Dimension Contrastive}
\cite{garrido2022duality} categorizes SSL methods as dimension-contrastive or sample-contrastive. Sample contrastive methods work by projecting sample augmentations (positives) closer to each other than that of other samples in a batch (negatives) \cite{chen2020simple}. This process causes samples to spread uniformly across the surface of a hypersphere \cite{wang2020understanding}. Other methods are derived from simple alterations to the definition of positive and negative sets. Research directions include using a momentum queue \cite{chen2020improved}, using nearest neighbors as positives \cite{dwibedi2021little}, enforcing cluster assignments \cite{caron2020unsupervised}, and enforcing hierarchical structures \cite{kokilepersaud2024taxes,kokilepersaud2024hex}. Dimension contrastive approaches enforce feature decorrelation through various methods. Examples include regularizing the embedding covariance matrix \cite{bardes2021vicreg,zbontar2021barlow,ermolov2021whitening} or introducing architectural constraints \cite{chen2021exploring,caron2021emerging} that implicitly regularizes dimensions. We also note that \cite{garrido2022duality} discussed conditions where both approaches are equivalent.  More recent work \cite{shwartz2023information} identified divergent entropy distributions between both approaches and suggested the development of methods that leverage both during training. Therefore, the novelty of our method can be understood as developing the first SSL method that identifies dynamics where both families of approaches should be integrated into the optimization process.

\paragraph{Understanding SSL Training Dynamics}
Work has also attempted to understand the training dynamics of SSL methods. \cite{jing2021understanding} analyzed the dimensional collapse phenomenon within contrastive learning settings. \cite{simon2023stepwise} characterized SSL training dynamics in terms of learning one eigenvalue at a time. \cite{tian2021understanding,srinath2023implicit} analyzed the learning dynamics of dimension contrastive methods in the context of linear networks. In general, work on understanding the underlying dynamics of SSL methods is limited.  Our work introduces SSL training dynamics based on the relationship between $I(R;Z)$ and $H(R)$.
\section{Analysis of Dimensionality and Mutual Information Relationship}
\label{sec:theory}

In this section, we address the following relevant research questions (\textbf{RQ}):

\begin{enumerate}
    \item \textbf{RQ.1} How do different optimization techniques influence the relationship between a representation space $R$ and its projection $Z$?
    \item \textbf{RQ.2}  How do  dynamics between $H(R)$ and $I(R;Z)$ manifest during standard SSL training?
    \item \textbf{RQ.3}   How does the relationship between H(R) and I(R;Z)  influence the downstream performance of the model?
\end{enumerate}
% How do different dimensionality optimization techniques (feature decorrelation vs. sample uniformity) influence the resulting relationship between $H(R)$ and $I(R;Z)$ at different stages of training?
% How do these dynamics between $H(R)$ and $I(R;Z)$ manifest in practical 
% How does the relationship between H(R) and I(R;Z) reached at the end of trianing influence the downstream perofrmance of the model?
\subsection{RQ.1 Optimization Impact on $R$ and $Z$}
\label{sec:dynamics}

 As discussed previously, SSL optimization involves increasing dimensionality through feature decorrelation or through sample uniformity. However, it is unknown how each type of optimization technique impacts the relationship between $R$ and $Z$ at different stages of training. To investigate this relationship, we perform a simulation within a Gaussian setting. Note that a gaussian assumption on the data space is standard practice in the analysis of SSL methods \cite{shwartz2023information}.  Assume that Gaussian distributed data is represented by $R \sim \mathcal{N}(\mu_{R},\Sigma_{R})$ where $R \in \mathcal{R}^{m}$. Additionally, assume that there is some projection of $R$ represented by $Z \sim \mathcal{N}(\mu_{Z},\Sigma_{Z})$ where $Z \in \mathcal{R}^{n}$ such that $n < m$. $R$ and $Z$ form a jointly multivariate normal distribution. Together, this distribution is defined by a block covariance matrix of the form $\Sigma = \begin{bmatrix}
    \Sigma_{Z} & \Sigma_{ZR} \\
    \Sigma_{RZ} & \Sigma_{R} \\
  \end{bmatrix}$ where $\Sigma_{ZR}$ and $\Sigma_{RZ}$ are the covariance matrices of both random variables. In this setting, the closed form solution for $I(R;Z) = \frac{1}{2}(ln(|\Sigma_R|) + ln(|\Sigma_Z|) - ln(|\Sigma|)).$ Applying the Schur complement to the block covariance matrix results in the following equation when all covariance matrices are invertible:

\begin{equation}
I(R;Z) = \frac{1}{2}(ln(|\Sigma_Z|) - ln(|Var(Z|R)|)) = \frac{1}{2}(ln(|\Sigma_R|) - ln(|Var(R|Z)|))
\label{eq:gauss_mutual}
\end{equation}

\begin{figure}[h!]
    \centering
    \includegraphics[width=\linewidth]{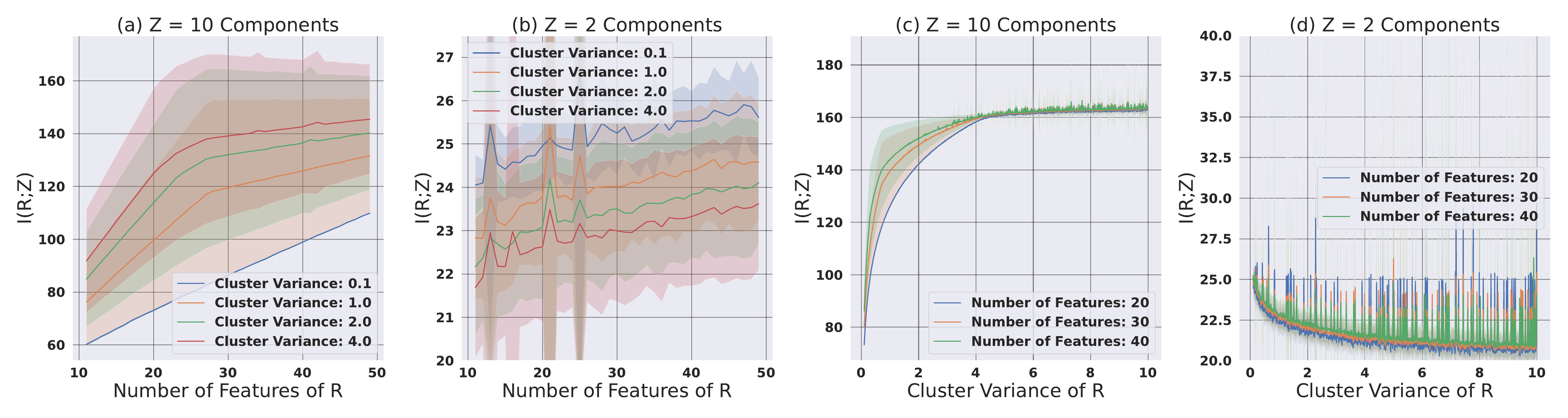}
    \caption{a) and b) show how the $I(R;Z)$ changes as the number of features of $R$ is increased. c) and d) show how $I(R;Z)$ varies as the sample cluster variance increases. Both pairs of Figures show the underlying dynamics under a large projection $Z=2$ and lesser projection $Z=10$ PCA components.}
    \label{fig:simulation}
\end{figure}

In equation \ref{eq:gauss_mutual}, $Var(Z|R) = \Sigma_{Z} - \Sigma_{ZR}\Sigma_{R}^{-1}\Sigma_{RZ}$ and $Var(R|Z) = \Sigma_{R} - \Sigma_{RZ}\Sigma_{Z}^{-1}\Sigma_{ZR}$. The details of this derivation can be found in Section \ref{sec:gaussian_derivation}. From this construction of the problem, several trends emerge. $I(R;Z)$ will increase or decrease depending on the relationship that the projection produces between $R$ and $Z$. Specifically, $I(R;Z)$ will increase when the variance of the space of interest increases while its corresponding conditional variance remains relatively lower. These variance changes can occur through a larger number of features or through a more uniform spread of data samples. Figure~\ref{fig:simulation} demonstrates a simulation of the effect of each by 
generating a synthetic gaussian dataset with 1000 samples, a defined variance for each of 5 generated clusters, and a defined number of features $m>10$ to simulate $R$.  $R$ is then projected with PCA to generate $Z$ with either 2 components or 10 components denoted by $n$.  Early in training, $R$ and $Z$ project closer to each other which is represented by the 10 component $Z$ space while later in training $R$ and $Z$ diverge to a greater degree represented by the 2 component projection. Additionally, PCA serves as a representative projection in this setting, since the information content of $R$ is  represented by the variance parameter of an $m$-dimensional Gaussian. The same simulation is repeated with similar conclusions in Section~\ref{sec:network_sim} where the projector is replaced with a small neural network. Further details can be found in Section \ref{sec:gaussian_sim}. Note that to avoid numerical instability due to a deterministic projection in Equation~\ref{eq:gauss_mutual}, we add a small amount of independent additive noise to the diagonal of each generated matrix.

In Figures~\ref{fig:simulation} a) and b), $H(R)$ increases by increasing the number of generated features $m$ while the cluster variance is kept constant. This corresponds to the feature decorrelation setting. The second experiment in Figures~\ref{fig:simulation} c) and d) involves increasing the sample variance of $R$ while keeping the number of features fixed which corresponds to the sample uniformity setting. In parts a) and b), for different cluster variance values, increasing the number of features in $R$ corresponds to an increase in $I(R;Z)$ regardless of the degree of projection. In parts c) and d), the behavior of $I(R;Z)$ varies significantly based on the degree of the projection. For the 10 component projection case of part c), increasing the sample variance initially increases $I(R;Z)$, but it gradually plateaus as the sample variance increases further. This suggests that the projection cannot capture the variance along certain dimensions after a specific point. In part d), in the 2 component case, increasing the sample variance by any amount reduces $I(R;Z)$. \textbf{Overall, this Figure shows that $I(R;Z)$ increases with a greater number of decorrelated features in $R$ regardless of the stage of training. In contrast, $I(R;Z)$ tends to plateau or decreases as sample uniformity increases later in training.} 

\subsection{RQ.2 Dynamics in Standard SSL Training}

\begin{wrapfigure}{l}{7.0cm}
    \centering
    \includegraphics[width=\linewidth]{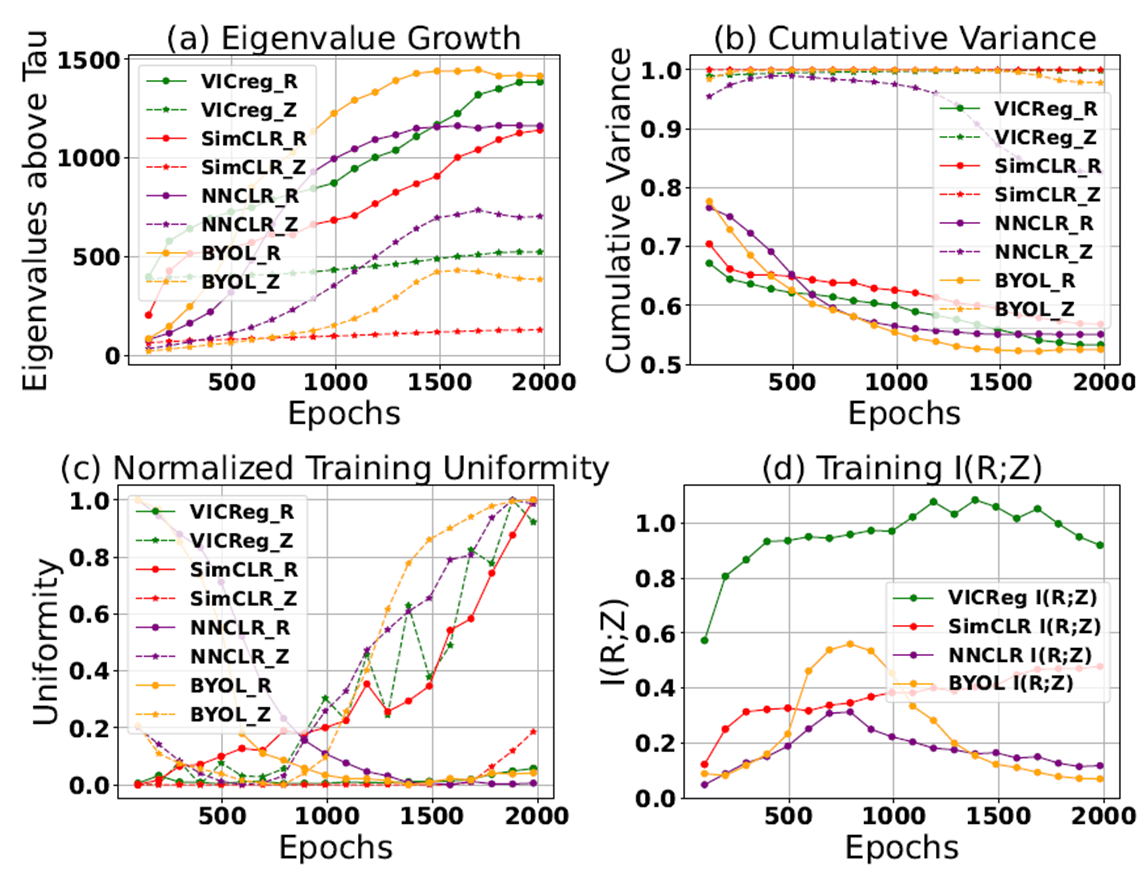}
    \caption{This figure analyzes the $R$ and $Z$ space for 4 different SSL models trained on Cifar-100 with ResNet-50. This analysis includes a)  the number of eigenvalues above a threshold of $\tau = .01$, b) the cumulative explained variance ratio for the top 30\% of eigenvalues, c) the uniformity of each space, and d) $I(R;Z)$.}
    \label{fig:2000_eig_simulation}
\end{wrapfigure}

To verify the simulated dynamics discussed in Section~\ref{sec:dynamics}, an empirical analysis within a real SSL setting is shown in Figure \ref{fig:2000_eig_simulation}. This experiment involves training a ResNet-50 model \cite{he2016deep} with 4 different SSL methods for 2000 epochs on Cifar-100. The projector is designed such that $R$ and $Z$ both have 2048 features. For all analytical experiments, the matrix analyzed is formed by passing each sample from the test set into the encoder network and then concatenating all representations. Further details can be found in Section~\ref{sec:eig_analysis}. In part a), we analyze the evolution of feature decorrelation for both the $R$ and $Z$ space across training by performing a count of the number of eigenvalues above a threshold $\tau=.01$. It is interesting to note that for the $R$ space the number of eigenvalues consistently increases until late in training while the $Z$ space has a more pronounced plateauing behavior earlier in training. This shows the behavior that the overall dimension of both spaces diverges from each other during training. In part b), we analyze the uniformity of eigenvalues by measuring what percentage of the variance in the space of interest is represented by the top 30\% of eigenvalues. This is known as the cumulative explained variance ratio \cite{jolliffe2016principal}. 
We observe that the cumulative explained variance of $R$ for all methods decreases during training which indicates that $H(R)$ is increasing due to a more uniform spread of eigenvalues and will gradually depend more on sample uniformity as training progresses. However, in $Z$, this metric is near 1.0 for all epochs of training which means that most of the variance of $Z$ is contained within only a small number of top eigenvalues. This suggests that samples in $Z$ distribute uniformly along a restricted subset of dimensions which is in contrast to the behavior of space $R$ that tries to distribute uniformly on as many dimensions as possible. This discrepancy in sample uniformity can also be visualized in part c) with the uniformity metric \cite{wang2020understanding}. We observe that for all SSL methods the uniformity between both spaces diverges from each other as training progresses. This divergent behavior is further confirmed in part d), where $I(R;Z)$  increases at the start of training, but gradually decreases for every method later in training. \textbf{This analysis confirms that SSL training dynamics exhibits different phases in the relationship between $H(R)$ and $I(R;Z)$. This result also suggests that optimization should be adaptively tuned between both feature decorrelation and sample uniformity as dimensionality increases.}

\subsection{RQ.3 Downstream Performance for a given $H(R)$ and $I(R;Z)$}

Another important consideration is how downstream performance is influenced by the relationship between $H(R)$ and $I(R;Z)$ at the end of training. To model this, the SSL information flow can be described by:  $Y \rightarrow X \rightarrow R \rightarrow Z \rightarrow T$. $Y$ represents the semantic concept associated with the data $X$. $T$ represents the associated SSL task. The end goal of the SSL objective is to maximize $I(Y;R)$ which is the mutual information between the semantics of the data and the representation space.  Recent work \cite{ouyang2025projection} showed that this setup results in an upper bound on $I(Y;R)$:

\begin{align}
    I(Y;R) \leq I(Y;Z) - I(R;Z) + H(R)
    \label{eq:upper_bound}
\end{align}

Our objective is to show how this bound is effected by the training dynamics discussed in Section~\ref{sec:dynamics} and to show that simply reducing $I(R;Z)$ and increasing $H(R)$ to maximize this bound is not possible given these dynamics. It is assumed that $R$ and $Z$ are drawn from a joint multivariate Gaussian distribution. Furthermore, $I(Y;Z)$ is assumed to approach some constant $G$ to isolate the analysis with respect to $I(R;Z)$ and $H(R)$. The justification for this term acting as a constant is from previous analyses \cite{saunshi2019theoretical} that assumed the information shared between semantic labels and the target SSL task can be regarded as a constant. Equation~\ref{eq:upper_bound} can then be rewritten as:

\begin{equation}
    I(Y;R) \leq G + \underbrace{\frac{1}{2}(ln(|\Sigma_R|)-ln(|\Sigma_Z|))}_{K(Both)} + \underbrace{\frac{1}{2}ln(|Var(Z|R)|)}_{V(I(R;Z))} + \underbrace{\frac{m}{2}(ln(2\pi)+1)}_{D(H(R))}
    \label{eq:info_gaussian}
\end{equation}
Equation~\ref{eq:info_gaussian} suggests that the bound on $I(Y;R)$ can be decomposed into three terms: a variance differential term $K$, a conditional variance term $V$, and a total dimension term $D$. Each term is labeled by its effect on $I(R;Z)$ or $H(R)$. Ideally, increasing each of these terms together would result in a higher overall bound on $I(Y;R).$ However, the SSL training dynamics discussed in Section~\ref{sec:dynamics} leads to the emergence of a dynamical system where increasing one of these terms can potentially limit the growth of the others. \textbf{This oscillatory behavior between each of these terms suggests that downstream performance $I(Y;R)$ depends on achieving an appropriate balance between all terms.}  The derivation of this bound, the exact details regarding the interactions between each term, and the assumptions needed for this analysis in a Gaussian context are discussed in Section~\ref{sec:info_theory}.

\begin{wrapfigure}{l}{7.0cm}
    \centering
    \includegraphics[width=\linewidth]{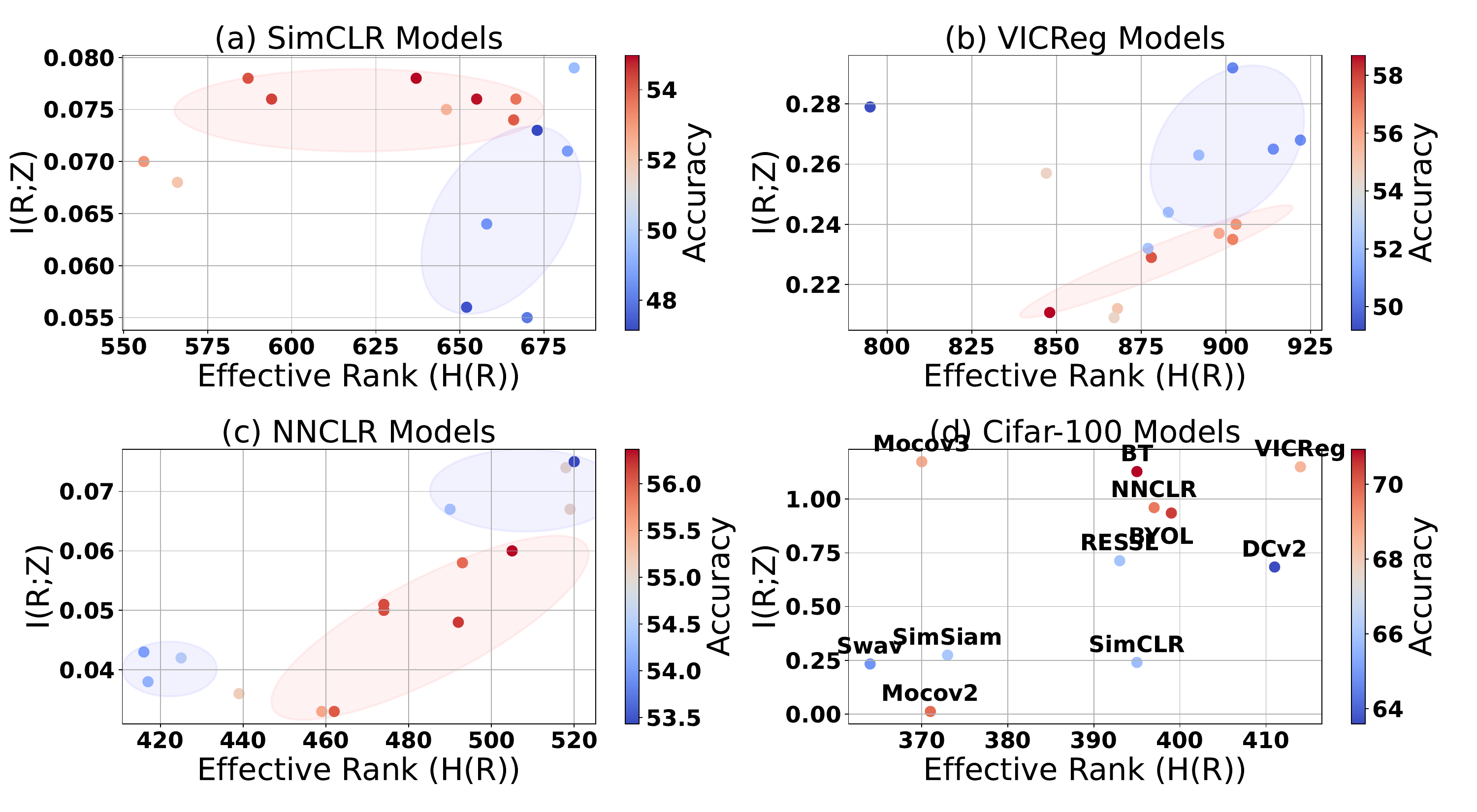}
    \caption{In Figures a), b), and c), the $H(R)$ and $I(R;Z)$ across 15 ResNet-50 models trained  with randomized hyperparameters with 3 different SSL strategies are shown. In Figure d), we show the same plot across 11 different SSL methods trained on ResNet-18.}
    \label{fig:random}
\end{wrapfigure}

 To empirically verify this behavior, we show how the relationship between $H(R)$ and $I(R;Z)$ impacts the downstream performance in Figure \ref{fig:random}. In parts a), b), and c) we train 15 different models with randomized hyperparameters specific to 3 different SSL methods with a ResNet-50 model on Cifar-100 for 100 epochs each. We observe that for each method, the best performing models cluster around specific $H(R)$ and $I(R;Z)$ values. This trend also holds in part d), where all 11 models are trained with entirely different SSL approaches. \textbf{This analysis suggests that SSL optimization should balance between both $H(R)$ and $I(R;Z)$ based on the discussed dynamics.}

\section{Methodology}

\begin{wrapfigure}{l}{5.5cm}
    \centering
    \includegraphics[width=\textwidth]{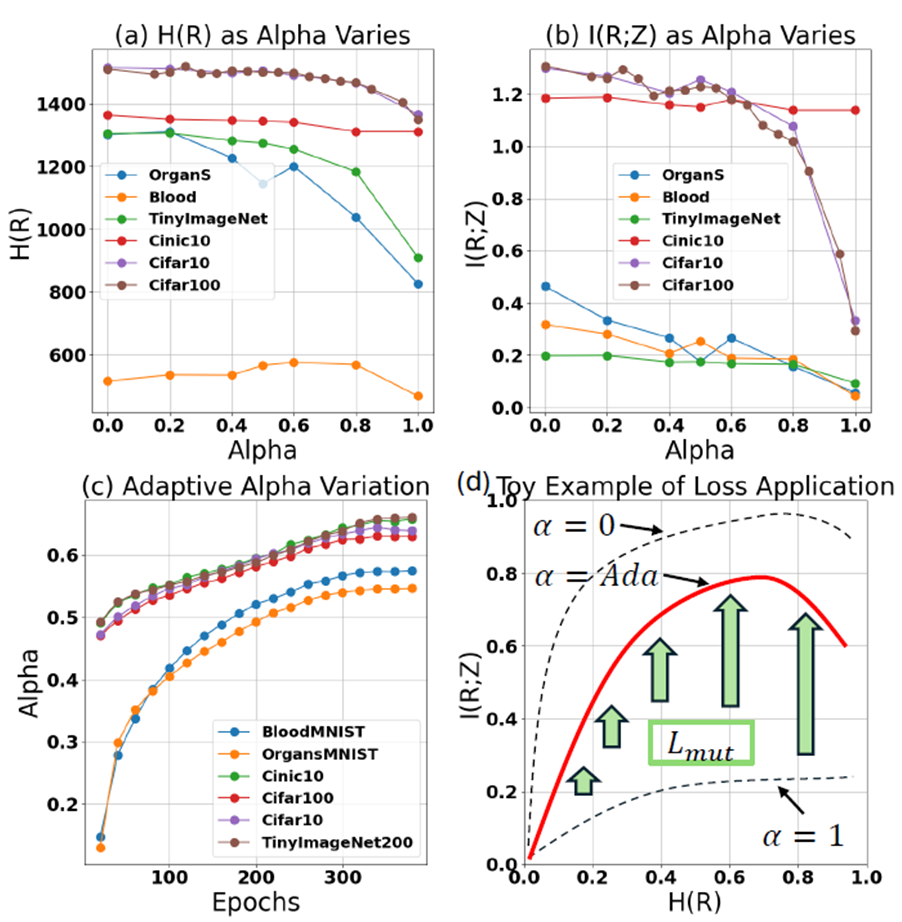}
    \caption{This figure shows the impact of manually varying alpha on a) $H(R)$ and b) $I(R;Z)$ in the setting where $\beta$ is set as a constant 0. c) This figure shows how the adaptive $\alpha$ parameter varies during training of a ResNet50 model for 400 epochs. d) This figure gives a toy example of the intuition behind our loss that includes the adaptive $\alpha$ alongside gradual $L_{mut}$ regularization.}
    \label{fig:alpha_study}
\end{wrapfigure}

 To achieve this balance, the loss needs different components that both support and oppose the growth of $H(R)$ and $I(R;Z)$ at different points of training by exploiting the observed dynamics discussed in Section~\ref{sec:theory}. The first set of components that are balanced with a $\alpha$ term is a loss based on sample uniformity $L_{SU}$ \cite{chen2020simple} and a loss based on feature decorrelation $L_{FD}$ \cite{bardes2021vicreg} that results in the total loss $\alpha L_{SU} + (1-\alpha)L_{FD}$. In parts a) and b) of Figure~\ref{fig:alpha_study}, we show the impact on $H(R)$ and $I(R;Z)$ when manually varying $\alpha$ from 0 to 1. This leads to the  behavior of parts a) and b), where gradually varying the loss from 0 ($L_{FD}$) to 1 ($L_{SU}$) consistently leads to lower $H(R)$ and $I(R;Z)$. However, we want to control the transition from feature decorrelation to sample uniformity optimization as the dimensionality increases. Therefore, we define $\alpha$ as an adaptive term that scales as a function of the dimensionality of the representation space: $\alpha = \frac{H(R)}{D}$. $H(R)$ is the effective rank of the training set representation space and $D$ represents the maximum effective rank possible for a given training set matrix. This growth of $\alpha$ across training for different data sets is shown in part c) of Figure~\ref{fig:alpha_study}. Therefore, the adaptive $\alpha$ term encourages an intermediate $H(R)$ and $I(R;Z)$ training trajectory when compared with $\alpha=0$ or $\alpha=1$ as shown in the toy intuition example of part d). However, at the end of training, both $L_{FD}$ and $L_{SU}$ will demonstrate the SSL dynamic of lowering $I(R;Z)$ in late stage training. Therefore, to maintain balance in this dynamic system, we need an additional term that explicitly opposes the decrease in $I(R;Z)$ as shown by the magnitude of $L_{mut}$ increasing as it scales with $\alpha$ in part d).

The above intuition results in our AdaDim algorithm. Consider an image $i$ drawn from a training pool $i \in I$. $i$ is passed into two random transformations $a(i) = x_i$ and $a^{'}(i) = x_i^{'}$ where $a$ and $a^{'}$ are drawn from the set of all random augmentations $A$. Both $x_i$ and $x_i^{'}$ are passed into an encoder network $e(\cdot)$. This results in the representations $e(x) = r_i$ and $e(x^{'})  = r_i^{'}$. These representations are then passed into a projection head $g(\cdot)$ that produces the embeddings $g(x_i) = z_i$ and $g(x_i^{'}) = z_i^{'}$. The collection of all representations and embeddings within a batch of $b$ samples and its corresponding augmentations can be represented by the $R$, $R^{'}$, $Z$, and $Z^{'}$ matrices. In this case, all matrices are composed of $b$ vectors with $F$ features. From this setup, we can compute $L_{SU}$ for the sample uniformity loss described in \cite{chen2020simple} and the $L_{FD}$ feature decorrelation loss described in \cite{bardes2021vicreg}. Note for embeddings passed into $L_{SU}$ further normalization is applied to produce $\hat{z_i}$. $L_{SU}$ enforces sample uniformity by distancing the image of interest $\hat{z_i}$ away from all other samples in the batch of interest. $L_{FD}$ operates as a feature decorrelation loss by driving the covariance matrix towards an identity matrix. Details of each loss are provided in Section \ref{sec:sim_vicreg}. We then compute the dimensionality of the current embedding space $Z$ after every $E_{\alpha}$ epochs (20 in this paper) of training. This is done by computing the SVD of 10 randomly chosen batches from the training set and then calculating the average effective rank across these batches $ER(Z)$ \cite{roy2007effective}. We then scale $ER(Z)$ by the maximum possible dimensionality value which is $D = min(2b,F)$ when we concatenate both $Z$ and $Z^{'}$ together to produce the adaptive parameter $\alpha = \frac{ER(Z)}{D}$. 

 \begin{wraptable}{r}{5.5cm}
\scalebox{.45}{
\begin{tabular}{ccccc}
\hline
\multicolumn{5}{c}{AdaDim Parameter Variation Ablation} \\ \hline
Method & $\alpha$       & $\gamma$ & $\beta$         & Accuracy                  \\ \hline
AdaDim & 0           & 0     & 0        & 72.14                     \\
AdaDim & 1           & 0     & 0        & 69.57                     \\
AdaDim & 0.5         & 0     & 0        & 72.23                     \\
AdaDim & Ada         & 0     & 0        & 72.30                      \\
AdaDim & Cosine      & 0     & 0        & 71.40                      \\
AdaDim & Linear      & 0     & 0        & 71.59                     \\
AdaDim & Ada         & 1e-04 & 1     & 72.00                        \\
AdaDim & 1           & 1e-04 & 1     & 68.81                     \\
AdaDim & 0           & 1e-04 & 1     & 72.10                      \\
AdaDim & Ada         & 1e-04   & Ada        & \textbf{72.73}            \\ \midrule
SimCLR + $\lambda$ \cite{ouyang2025projection} & -  & - & -  &  69.11\\
VICReg + $\lambda$ \cite{ouyang2025projection} & -  & - & - & 72.14 \\ \hline

\end{tabular}}
    \caption{This table studies the effect of variations to the $\alpha$, $\beta$, and $\gamma$ parameters on performance. Experiments use a ResNet-50 model on Cifar-100 for 400 epochs with the baseline hyperparameter setting.}\label{tab:parameter_var}
\end{wraptable}

We also want to gradually increase regularization on $I(R;Z)$ to counter the decrease in $I(R;Z)$ that emerges later in SSL training. To do this, we compute an $I(R;Z)$ loss $L_{mut}(R,Z)$ that encourages higher $I(R;Z)$ with the $\alpha$-Renyi entropy approximation technique \cite{zhang2023matrix,ouyang2025projection,renyi1961measures}. This loss first computes the entropy of a matrix with the formula $H(R) = -\frac{1}{2}log[tr(\frac{R}{b})^2]$. The mutual information can then be computed as $I(R;Z) = H(R) + H(Z) - H(R\odot Z)$. For the purpose of numerical stability, the $I(R;Z)$ loss is computed as $L_{mut} = I(\hat{R}\hat{R}^T;\hat{Z^{'}}\hat{Z^{'}}^T)$ where $\hat{R}$ and $\hat{Z^{'}}$ refer to the normalized version of each space. We scale its regularization through the term $\beta = \gamma * \alpha$ with $\gamma$ set as a constant. We provide PyTorch style pseudo-code of our method in Section~\ref{sec:pseudocode}. Our final loss is: 

 \begin{equation}
L_{AdaDim} = (1-\beta)[(1-\alpha) L_{FD} + \alpha L_{SU}] - \beta L_{mut}
\label{eq:AdaDim}
 \end{equation}

\begin{wraptable}{l}{8.5cm}
\centering
\scalebox{0.45}{
\begin{tabular}{@{}ccccccccc@{}}
\toprule
\multicolumn{9}{c}{AdaDim Diverse Dataset Comparison}                                                                                                                      \\ \midrule
Method                          & Cifar100                    & TinyImageNet200 & Cinic10 & STL10           & Blood          & OrganA         & OrganS         & OrganC         \\ \midrule
SimCLR  \cite{chen2020simple}                       & 69.06                       & 46.66           &   78.77      &     86.73            & 93.10          & 88.04          & 77.98          & 91.13          \\
VICReg   \cite{bardes2021vicreg}                       & 72.18                       & 48.47           &  82.70       &      87.92           & 93.77          & 92.21          & 80.37          & 91.84          \\
Moco v2     \cite{chen2020improved}                    & 71.01                       & 46.78     &   81.48      &      92.41           & 93.74 & 90.49          & 75.96          & 90.81          \\
BYOL     \cite{grill2020bootstrap}                       & 71.72                & 32.96           &  80.00       &       89.96         & 92.45          & 92.26          & 78.53          & 91.45          \\
Barlow Twins      \cite{bardes2021vicreg}              & 70.84                 & 46.73           &    81.5     &       88.45          & 89.91          & 91.69          & 78.69          & 89.77          \\
NNCLR       \cite{dwibedi2021little}                    & 70.72                       & 39.66           &  77.28       &        87.16         & 93.15          & 92.93 &  79.92    & 91.71          \\
SimSiam     \cite{chen2021exploring}                    & 65.52                       & 31.35           &  79.97       &     89.45            & 91.78          & 91.91          & 78.31          & 90.79          \\
Deepcluster v2     \cite{caron2018deep}             & 65.70                       & 41.87           &    74.80     &      82.93           & 93.56          & 92.21          & 77.93          & 74.31          \\
Moco v3      \cite{caron2021emerging}                   & 63.96                       & 37.56           &     74.71    &        85.25         & 93.33          & 92.27          & 78.69          & 91.84    \\ \midrule
AdaDim ($\gamma$ = 0)                         & 72.23              & 47.87  &     82.38    &        88.11         & 93.74 &  92.90    & 80.19 & \underline{91.95} \\ AdaDim ($\gamma=1e-4$)                         & \underline{72.73}              & 48.76  &  82.77      &       89.01          & \underline{94.24} &  92.77    & \textbf{80.80} & \underline{91.95} \\ 

 AdaDim ($\gamma$=Tuned)                         & \underline{72.73}              & 48.76  & 82.84      &       89.21          & \underline{94.24} &  \underline{93.34}    & \textbf{80.80} & \textbf{92.33} \\\midrule
 Resa \cite{weng2025clustering}                         & 72.06              & \underline{50.73}  &   \underline{84.23}      &       \underline{92.56}          & 88.71 &  93.20    & 78.81 & 91.50\\
 AdaDim + momentum                         & \textbf{75.38}              & \textbf{54.68}  &   \textbf{84.99}      &       \textbf{92.70}          & \textbf{95.21} &  \textbf{93.67}    & 78.83 & 91.82\\
 \bottomrule
\end{tabular}}
\caption{Comparison methods use the given parameters from \cite{da2022solo}. Experiments involve a ResNet-50 model for 400 epochs with baseline hyperparameters as well as comparisons with an additional momentum encoder and the Resa method \cite{weng2025clustering} under the expanded hyperparameter setting. For the AdaDim baseline setting, we vary $\gamma$ to 0, 1e-4, and a value tuned to specific datasets.}
\label{tab: dataset_wrapped}
\end{wraptable}

\section{Results}
\label{sec:results}

%For the vast majority of experiments, a Resnet-50 \cite{he2016deep} architecture is used in tandem  with a  simple 3-layer MLP projection head. All ablation study experiments utilize the same projection head with augmentation scheme for ease of comparison. For comparison with state of the art models, the parameters from the solo-learn \cite{da2022solo} library are used. Our AdaDim method is trained with a LARS optimizer, batch size of 256, a learning rate of 0.4, a weight decay of 1e-4, and $\gamma = 1e-4$. For all experiments, models are trained for 400 epochs. The exception to these conventions are comparisons with ImageNet-100 where a ResNet-18 model is used with $\gamma = -0.1$.  An online linear evaluation setting is used for all experiments that has been shown to directly correlate with the offline setting and act as a standard benchmark \cite{garrido2023rankme,garrido2022duality,da2022solo}. Further details are in Section~\ref{sec:method_details}.

To thoroughly analyze AdaDim, we introduce two different sets of hyperparameters: the \textbf{baseline and expanded settings}. In the baseline setting, we use the traditional joint embedding setup with a single ResNet-50 \cite{he2016deep} backbone in tandem with a simple 3-layer MLP projector that has an output dimension of 2048. The baseline setting also uses the LARS optimizer, batch size of 256, a learning rate of 0.4, temperature of 0.1, and a weight decay of 1e-4. We also use the asymmetric augmentation scheme of \cite{da2022solo}. Evaluation is performed with an online linear predictor that correlates with the offline setting \cite{garrido2022duality,garrido2023rankme}. In the expanded setting, we use the same AdaDim loss function, but utilize an additional momentum encoder. For these experiments, we use the training hyperparameters of the state of the art Resa algorithm \cite{weng2025clustering}. Evaluation is performed in the offline manner of \cite{ermolov2021whitening}. All experiments generally use 20 epochs between every $\alpha$ update and a default gamma of $1e-4$. Further details are in Section~\ref{sec:method_details}. %\input{iclr2026/Tables/resnet_sota_comparison}

%\input{iclr2026/Tables/sota_table_ablation_together}

%\input{iclr2026/Tables/ssl_sota_comparison_table}

%\input{iclr2026/Tables/ICLR_Revision/solo_learn_table}

\begin{comment}
\begin{figure}[h!]
    \centering
    \includegraphics[width =\columnwidth]{Figures/16_different_epochs.pdf}
    \caption{a) This figure demonstrates the performance of AdaDim under different amounts of training epochs on Cifar-100. b) This figure shows how AdaDim performs under a long training setting  for Cifar-10 (C10), Cifar-100 (C100), and TinyImageNet200 (TIN200) c) This shows the $H(R)$ and $I(R;Z)$ trajectory over 2000 epochs on 4 different algorithms.}
    \label{fig:training_epochs}
\end{figure}
\end{comment}

\begin{wraptable}{r}{6.5cm}
\scalebox{0.45}{
\begin{tabular}{cccccccc}
\hline
\multicolumn{8}{c}{Solo-Learn \cite{da2022solo} Benchmark Comparison}                        \\ \hline
Method          & M & P & Q & C & Cifar-10       & Cifar-100      & ImageNet-100   \\ \hline
Barlow Twins \cite{zbontar2021barlow}    & \xmark & \xmark & \xmark & \xmark & 92.10          & 70.90          & 80.38          \\
BYOL \cite{grill2020bootstrap}            & \checkmark & \checkmark & \xmark & \xmark & 92.58          & 70.46          & 80.16          \\
Deep Cluster v2 \cite{caron2018deep} & \xmark & \xmark & \xmark & \checkmark & 88.85          & 63.61          & 80.32          \\
DINO     \cite{caron2021emerging}       & \checkmark & \xmark & \xmark & \xmark& 89.52          & 66.76          & 74.84          \\
Moco v2  \cite{chen2020improved}       & \checkmark & \xmark & \checkmark & \xmark& 92.94          & 69.89          & 78.20          \\
Moco v3 \cite{chen2021empirical}        & \checkmark & \checkmark & \xmark & \xmark & 93.10          & 68.83          & 80.36          \\
NNCLR   \cite{dwibedi2021little}        & \xmark & \checkmark & \checkmark & \checkmark& 91.88          & 69.62          & 79.80          \\
ReSSL    \cite{zheng2021ressl}       & \checkmark & \xmark & \checkmark & \xmark& 90.63          & 65.92          & 76.92          \\
SimCLR   \cite{chen2020simple}       & \xmark & \xmark & \xmark & \xmark & 90.74          & 65.78          & 77.64          \\
SimSiam   \cite{chen2021exploring}      & \xmark & \checkmark & \xmark & \xmark & 90.51          & 66.04          & 74.54          \\
SwAV  \cite{caron2020unsupervised}          & \xmark & \xmark & \xmark & \checkmark & 89.17          & 64.88          & 74.04          \\
VICReg   \cite{bardes2021vicreg}       & \xmark & \xmark & \xmark & \xmark & 92.07          & 68.54          & 79.22          \\
AdaDim (Ours)         & \xmark & \xmark & \xmark & \xmark & 92.81          & 71.20          & 80.78          \\ \hline
Resa    \cite{weng2025clustering}        & \checkmark & \xmark & \xmark & \checkmark & \underline{93.53} &  \underline{72.21}    &  \underline{82.24}    \\
AdaDim (Ours)         & \checkmark & \xmark & \xmark & \xmark & \textbf{93.61}    & \textbf{74.31} & \textbf{83.10} \\ \hline
\end{tabular}}
\caption{This table is a solo-learn benchmark comparison with all methods using a ResNet-18 model trained for 1000 epochs on Cifar-10 and Cifar-100 with a batch size of 256. ImageNet-100 experiments are trained for 400 epochs with a batch size of 128. (M = Momentum, P = Predictor, Q = Queue, C = Clustering)}\label{tab:solo_learn_wrapped}
\end{wraptable}

 In Table~\ref{tab:parameter_var}, we analyze the performance impact of different hypothetical design choices on $\alpha$, $\gamma$, and $\beta$. Note that while AdaDim only introduces a single hyperparameter $\gamma$, we individually tune each in this table to validate the adaptive nature of our method. To start, we compare against methods that make use of heuristic $\alpha$ scaling methods without an adaptive computation such as cosine or linear schedules between 0 and 1 over the course of training. These methods underperform relative to the adaptive case and highlights the importance of adapting the optimization based on the dimensional characteristics of the dataset. We also compare against using a fixed $\gamma$ term during training. We observe that this regularization causes a slight decrease in performance. This result suggests that $I(R;Z)$ regularization should be applied selectively at specific points in SSL training, rather than as a constant term throughout. This intuition is further confirmed by comparing against the $\lambda$ regularized $I(R;Z)$ reduction loss proposed in \cite{ouyang2025projection}. 
 This work argues that simply reducing $I(R;Z)$ during training without an adaptive mechanism can improve SSL representations. However, simply reducing $I(R;Z)$ doesn't conform with the dynamics we discuss where $I(R;Z)$ goes through periods of growth and reduction during training. 
 Consequently, we find that using their suggested $\lambda$ parameter results in little to no improvement in our baseline setting. The reason for this discrepancy with their results may be that in their original paper their method only showed improvements with 200 epochs of training. In this limited setting, fixed regularization may work as there is limited training time for the discussed dynamics to emerge. We further validate these findings across more datasets and a fixed parameter setting in Section~\ref{sec:alternate_strategies}. We also note a performance improvement when transitioning between training with $\alpha$ alone compared to $\beta$ in tandem with $\alpha$.  This result confirms the importance of balancing between both $H(R)$ and $I(R;Z)$ regularization in a manner consistent with the discussed training dynamics. We confirm this ablation study across many different datasets in Table~\ref{tab: dataset_wrapped} and in comparison with state of the art approaches. We find that our method out performs or is comparable to other approaches across diverse settings with a fixed $\gamma$ choice of $1e-4$. Additionally, further performance improvements can be attained by tuning $\gamma$ with respect to each dataset individually which is a result of better tuning towards the different dimensionality characteristics of each dataset.
 
  \begin{wraptable}{l}{5.5cm}
\scalebox{0.4}{
\begin{tabular}{cccccc}
\hline
\multicolumn{4}{c}{ImageNet SOTA Comparison} \\ \hline
Method        & Batch Size    & R50(100) & R50(200) & ViT-S(100)    \\ \hline
SimCLR  \cite{chen2020simple}      & 4096          & 66.50     & 68.30 & -   \\
SwAV    \cite{zhan2020online}       & 4096          & 66.50  & 69.10 &  -    \\
Moco v3  \cite{chen2021empirical}     & 1024          & 67.40  & 71.00 &   -   \\
BYOL     \cite{grill2020bootstrap}    & 4096          & 66.50  & 70.60 &   -   \\
Barlow Twins \cite{zbontar2021barlow}   & 2048          & 67.70   &70.20 &   -   \\
VICReg  \cite{bardes2021vicreg}     & 2048          & 68.60   & 70.80 &  -   \\
SimSiam  \cite{chen2021exploring}    & 256           & 68.10   & 70.0 &   -  \\
INTL   \cite{weng2023modulate}   & 1024          & 69.70   & 71.20 &    -  \\
MEC    \cite{liu2022self}    & 1024          & 70.60   & 71.90 &    - \\  \midrule
Resa   \cite{weng2025clustering}  & 256           &  70.80    & 71.60 &   67.95     \\
AdaDim (Ours)       & 256           & \textbf{71.01} & \textbf{72.84} &  \textbf{68.31}   \\ 
\hline
\end{tabular}}
  \caption{This table shows a state of the art ImageNet comparison within the single crop setting on ResNet-50 (R50) for 100 and 200 epochs and ViT-Small (ViT-S) for 100 Epochs. All comparison results are taken from their original papers or tables in \cite{weng2025clustering}. For comparisons with Resa, we run their algorithm with their provided codebase and architectural settings.}
  \label{tab:imagenet_wrapped}
\end{wraptable}

 The ability of AdaDim to adapt to unique data settings is also illustrated with consistent performance improvements on the MedMNIST datasets \cite{yang2023medmnist} that exhibits characteristics that SSL algorithms traditionally struggle with \cite{cole2022does}. Furthermore, adding a momentum encoder as part of the AdaDim methodology leads to significant performance improvements for the majority of datasets. This emphasizes that AdaDim can effectively leverage the architectural techniques of other methods. We also compare our method in the expanded hyperparameter setting in Tables~\ref{tab:solo_learn_wrapped} and ~\ref{tab:imagenet_wrapped} on standardized benchmarks \cite{da2022solo,deng2009imagenet}.  We specify whether each comparison method utilizes additional SSL techniques such as a momentum encoder(M), predictor head (P), queue (Q), or clustering technique (C).  In Table~\ref{tab:solo_learn_wrapped}, we note that without using any of these techniques, we are able to achieve performance that out-performs other methods with the only additional overhead being an SVD calculation on 10 batches every 20 epochs. We validate the low compute cost of our method in Section~\ref{sec:compute_method}. However, we also show that our method makes use of a momentum encoder particularly well to out-perform all existing methods by a significant margin. 
 
  \begin{wrapfigure}{r}{5.5cm}
    \centering
    \includegraphics[width =\columnwidth]{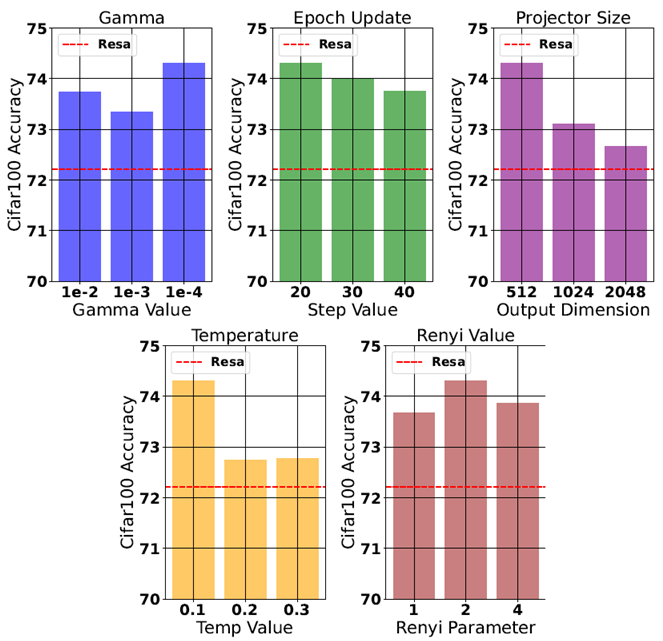}
    \caption{We show the impact of different hyperparameters on performance with 1000 epochs of ResNet-18 training on Cifar-100. Experiments are performed in the expanded hyperparameter setting. We also indicate the Resa \cite{weng2025clustering} baseline performance.}
    \label{fig:hyparameter_var_2}
\end{wrapfigure}
 
This improvement may be due to the momentum encoder acting as a stable update of the representation space thereby allowing AdaDim to better leverage training dynamics. We also observe in Table ~\ref{tab:imagenet_wrapped}  similar improvements over baselines on the large scale ImageNet dataset where AdaDim out-performs the most recent state of the art Resa algorithm \cite{weng2025clustering} without requiring an additional clustering step on both ResNet-50 and ViT-Small architectures. Further training to 200 epochs also increases the margin of performance improvement of AdaDim compared to all other methods. This suggests that our method benefits more from longer training that requires an adaptation to dynamics. This trend holds across different datasets and amount of training in Figure~\ref{fig:training_epochs}.  In Figure~\ref{fig:hyparameter_var_2}, we also find that these improvements over \cite{weng2025clustering} are maintained with variations in the parameter choices made. Examples of parameters that we alter include $\gamma$, $E_{\alpha}$, the projector size, and  the $L_{SU}$ temperature parameter. Overall, these experiments highlight the robustness of our method.

\section{Conclusion}
This paper demonstrates that the best performing SSL models arrive at a balance between the dimensionality $H(R)$ of the representation space and the mutual information between the representation and embedding spaces $I(R;Z)$. Additionally, we find that this balance emerges through specific dynamics that show that increases in $H(R)$ due to feature decorrelation are preserved between $R$ and $Z$, but increases due to samples  spreading  uniformly can cause $I(R;Z)$ to increase, plateau, or decrease depending different phases of training. These results are significant in  SSL literature, as they demonstrate that both families of SSL losses should be applied adaptively at different points during training. We introduce a method called AdaDim that takes advantage of these insights to achieve state of the art performance. Limitations of our work are discussed in Section~\ref{sec:limit}.

\bibliography{bib}

\begin{thebibliography}{56}
\providecommand{\natexlab}[1]{#1}
\providecommand{\url}[1]{\texttt{#1}}
\expandafter\ifx\csname urlstyle\endcsname\relax
  \providecommand{\doi}[1]{doi: #1}\else
  \providecommand{\doi}{doi: \begingroup \urlstyle{rm}\Url}\fi

\bibitem[Agrawal et~al.(2022)Agrawal, Mondal, Ghosh, and Richards]{agrawal2022alpha}
Agrawal, K.~K., Mondal, A.~K., Ghosh, A., and Richards, B.
\newblock alpha-req: Assessing representation quality in self-supervised learning by measuring eigenspectrum decay.
\newblock \emph{Advances in Neural Information Processing Systems}, 35:\penalty0 17626--17638, 2022.

\bibitem[Bardes et~al.(2021)Bardes, Ponce, and LeCun]{bardes2021vicreg}
Bardes, A., Ponce, J., and LeCun, Y.
\newblock Vicreg: Variance-invariance-covariance regularization for self-supervised learning.
\newblock \emph{arXiv preprint arXiv:2105.04906}, 2021.

\bibitem[Caron et~al.(2018)Caron, Bojanowski, Joulin, and Douze]{caron2018deep}
Caron, M., Bojanowski, P., Joulin, A., and Douze, M.
\newblock Deep clustering for unsupervised learning of visual features.
\newblock In \emph{Proceedings of the European conference on computer vision (ECCV)}, pp.\  132--149, 2018.

\bibitem[Caron et~al.(2020)Caron, Misra, Mairal, Goyal, Bojanowski, and Joulin]{caron2020unsupervised}
Caron, M., Misra, I., Mairal, J., Goyal, P., Bojanowski, P., and Joulin, A.
\newblock Unsupervised learning of visual features by contrasting cluster assignments.
\newblock \emph{Advances in neural information processing systems}, 33:\penalty0 9912--9924, 2020.

\bibitem[Caron et~al.(2021)Caron, Touvron, Misra, J{\'e}gou, Mairal, Bojanowski, and Joulin]{caron2021emerging}
Caron, M., Touvron, H., Misra, I., J{\'e}gou, H., Mairal, J., Bojanowski, P., and Joulin, A.
\newblock Emerging properties in self-supervised vision transformers.
\newblock In \emph{Proceedings of the IEEE/CVF international conference on computer vision}, pp.\  9650--9660, 2021.

\bibitem[Chen et~al.(2020{\natexlab{a}})Chen, Kornblith, Norouzi, and Hinton]{chen2020simple}
Chen, T., Kornblith, S., Norouzi, M., and Hinton, G.
\newblock A simple framework for contrastive learning of visual representations.
\newblock In \emph{International conference on machine learning}, pp.\  1597--1607. PMLR, 2020{\natexlab{a}}.

\bibitem[Chen \& He(2021)Chen and He]{chen2021exploring}
Chen, X. and He, K.
\newblock Exploring simple siamese representation learning.
\newblock In \emph{Proceedings of the IEEE/CVF conference on computer vision and pattern recognition}, pp.\  15750--15758, 2021.

\bibitem[Chen et~al.(2020{\natexlab{b}})Chen, Fan, Girshick, and He]{chen2020improved}
Chen, X., Fan, H., Girshick, R., and He, K.
\newblock Improved baselines with momentum contrastive learning.
\newblock \emph{arXiv preprint arXiv:2003.04297}, 2020{\natexlab{b}}.

\bibitem[Chen et~al.(2021)Chen, Xie, and He]{chen2021empirical}
Chen, X., Xie, S., and He, K.
\newblock An empirical study of training self-supervised vision transformers.
\newblock In \emph{Proceedings of the IEEE/CVF international conference on computer vision}, pp.\  9640--9649, 2021.

\bibitem[Cole et~al.(2022)Cole, Yang, Wilber, Mac~Aodha, and Belongie]{cole2022does}
Cole, E., Yang, X., Wilber, K., Mac~Aodha, O., and Belongie, S.
\newblock When does contrastive visual representation learning work?
\newblock In \emph{Proceedings of the IEEE/CVF conference on computer vision and pattern recognition}, pp.\  14755--14764, 2022.

\bibitem[Da~Costa et~al.(2022)Da~Costa, Fini, Nabi, Sebe, and Ricci]{da2022solo}
Da~Costa, V. G.~T., Fini, E., Nabi, M., Sebe, N., and Ricci, E.
\newblock solo-learn: A library of self-supervised methods for visual representation learning.
\newblock \emph{Journal of Machine Learning Research}, 23\penalty0 (56):\penalty0 1--6, 2022.

\bibitem[Darlow et~al.(2018)Darlow, Crowley, Antoniou, and Storkey]{darlow2018cinic}
Darlow, L.~N., Crowley, E.~J., Antoniou, A., and Storkey, A.~J.
\newblock Cinic-10 is not imagenet or cifar-10.
\newblock \emph{arXiv preprint arXiv:1810.03505}, 2018.

\bibitem[Deng et~al.(2009)Deng, Dong, Socher, Li, Li, and Fei-Fei]{deng2009imagenet}
Deng, J., Dong, W., Socher, R., Li, L.-J., Li, K., and Fei-Fei, L.
\newblock Imagenet: A large-scale hierarchical image database.
\newblock In \emph{2009 IEEE conference on computer vision and pattern recognition}, pp.\  248--255. Ieee, 2009.

\bibitem[Dosovitskiy(2020)]{dosovitskiy2020image}
Dosovitskiy, A.
\newblock An image is worth 16x16 words: Transformers for image recognition at scale.
\newblock \emph{arXiv preprint arXiv:2010.11929}, 2020.

\bibitem[Dwibedi et~al.(2021)Dwibedi, Aytar, Tompson, Sermanet, and Zisserman]{dwibedi2021little}
Dwibedi, D., Aytar, Y., Tompson, J., Sermanet, P., and Zisserman, A.
\newblock With a little help from my friends: Nearest-neighbor contrastive learning of visual representations.
\newblock In \emph{Proceedings of the IEEE/CVF International Conference on Computer Vision}, pp.\  9588--9597, 2021.

\bibitem[Ermolov et~al.(2021)Ermolov, Siarohin, Sangineto, and Sebe]{ermolov2021whitening}
Ermolov, A., Siarohin, A., Sangineto, E., and Sebe, N.
\newblock Whitening for self-supervised representation learning.
\newblock In \emph{International conference on machine learning}, pp.\  3015--3024. PMLR, 2021.

\bibitem[Garrido et~al.(2022)Garrido, Chen, Bardes, Najman, and Lecun]{garrido2022duality}
Garrido, Q., Chen, Y., Bardes, A., Najman, L., and Lecun, Y.
\newblock On the duality between contrastive and non-contrastive self-supervised learning.
\newblock \emph{arXiv preprint arXiv:2206.02574}, 2022.

\bibitem[Garrido et~al.(2023)Garrido, Balestriero, Najman, and Lecun]{garrido2023rankme}
Garrido, Q., Balestriero, R., Najman, L., and Lecun, Y.
\newblock Rankme: Assessing the downstream performance of pretrained self-supervised representations by their rank.
\newblock In \emph{International conference on machine learning}, pp.\  10929--10974. PMLR, 2023.

\bibitem[Grill et~al.(2020)Grill, Strub, Altch{\'e}, Tallec, Richemond, Buchatskaya, Doersch, Avila~Pires, Guo, Gheshlaghi~Azar, et~al.]{grill2020bootstrap}
Grill, J.-B., Strub, F., Altch{\'e}, F., Tallec, C., Richemond, P., Buchatskaya, E., Doersch, C., Avila~Pires, B., Guo, Z., Gheshlaghi~Azar, M., et~al.
\newblock Bootstrap your own latent-a new approach to self-supervised learning.
\newblock \emph{Advances in neural information processing systems}, 33:\penalty0 21271--21284, 2020.

\bibitem[He et~al.(2016)He, Zhang, Ren, and Sun]{he2016deep}
He, K., Zhang, X., Ren, S., and Sun, J.
\newblock Deep residual learning for image recognition.
\newblock In \emph{Proceedings of the IEEE conference on computer vision and pattern recognition}, pp.\  770--778, 2016.

\bibitem[Jing et~al.(2021)Jing, Vincent, LeCun, and Tian]{jing2021understanding}
Jing, L., Vincent, P., LeCun, Y., and Tian, Y.
\newblock Understanding dimensional collapse in contrastive self-supervised learning.
\newblock \emph{arXiv preprint arXiv:2110.09348}, 2021.

\bibitem[Jolliffe \& Cadima(2016)Jolliffe and Cadima]{jolliffe2016principal}
Jolliffe, I.~T. and Cadima, J.
\newblock Principal component analysis: a review and recent developments.
\newblock \emph{Philosophical transactions of the royal society A: Mathematical, Physical and Engineering Sciences}, 374\penalty0 (2065):\penalty0 20150202, 2016.

\bibitem[Kim et~al.(2023)Kim, Kang, Hwang, Shin, and Rhee]{kim2023vne}
Kim, J., Kang, S., Hwang, D., Shin, J., and Rhee, W.
\newblock Vne: An effective method for improving deep representation by manipulating eigenvalue distribution.
\newblock In \emph{Proceedings of the IEEE/CVF Conference on Computer Vision and Pattern Recognition}, pp.\  3799--3810, 2023.

\bibitem[Kokilepersaud et~al.(2022)Kokilepersaud, Prabhushankar, and AlRegib]{kokilepersaud2022volumetric}
Kokilepersaud, K., Prabhushankar, M., and AlRegib, G.
\newblock Volumetric supervised contrastive learning for seismic semantic segmentation.
\newblock In \emph{Second International Meeting for Applied Geoscience \& Energy}, pp.\  1699--1703. Society of Exploration Geophysicists and American Association of Petroleum~…, 2022.

\bibitem[Kokilepersaud et~al.(2023{\natexlab{a}})Kokilepersaud, Corona, Prabhushankar, AlRegib, and Wykoff]{kokilepersaud2023clinically}
Kokilepersaud, K., Corona, S.~T., Prabhushankar, M., AlRegib, G., and Wykoff, C.
\newblock Clinically labeled contrastive learning for oct biomarker classification.
\newblock \emph{IEEE Journal of Biomedical and Health Informatics}, 27\penalty0 (9):\penalty0 4397--4408, 2023{\natexlab{a}}.

\bibitem[Kokilepersaud et~al.(2023{\natexlab{b}})Kokilepersaud, Prabhushankar, Yarici, AlRegib, and Parchami]{kokilepersaud2023exploiting}
Kokilepersaud, K., Prabhushankar, M., Yarici, Y., AlRegib, G., and Parchami, A.
\newblock Exploiting the distortion-semantic interaction in fisheye data.
\newblock \emph{IEEE Open Journal of Signal Processing}, 4:\penalty0 284--293, 2023{\natexlab{b}}.

\bibitem[Kokilepersaud et~al.(2024{\natexlab{a}})Kokilepersaud, Kim, Prabhushankar, and AlRegib]{kokilepersaud2024hex}
Kokilepersaud, K., Kim, S., Prabhushankar, M., and AlRegib, G.
\newblock Hex: Hierarchical emergence exploitation in self-supervised algorithms.
\newblock \emph{arXiv preprint arXiv:2410.23200}, 2024{\natexlab{a}}.

\bibitem[Kokilepersaud et~al.(2024{\natexlab{b}})Kokilepersaud, Yarici, Prabhushankar, and AlRegib]{kokilepersaud2024taxes}
Kokilepersaud, K., Yarici, Y., Prabhushankar, M., and AlRegib, G.
\newblock Taxes are all you need: Integration of taxonomical hierarchy relationships into the contrastive loss.
\newblock \emph{arXiv preprint arXiv:2406.06848}, 2024{\natexlab{b}}.

\bibitem[Krizhevsky et~al.(2009)Krizhevsky, Hinton, et~al.]{krizhevsky2009learning}
Krizhevsky, A., Hinton, G., et~al.
\newblock Learning multiple layers of features from tiny images.
\newblock 2009.

\bibitem[Liu et~al.(2022)Liu, Wang, Li, and Wang]{liu2022self}
Liu, X., Wang, Z., Li, Y.-L., and Wang, S.
\newblock Self-supervised learning via maximum entropy coding.
\newblock \emph{Advances in neural information processing systems}, 35:\penalty0 34091--34105, 2022.

\bibitem[Liu et~al.(2021)Liu, Lin, Cao, Hu, Wei, Zhang, Lin, and Guo]{liu2021swin}
Liu, Z., Lin, Y., Cao, Y., Hu, H., Wei, Y., Zhang, Z., Lin, S., and Guo, B.
\newblock Swin transformer: Hierarchical vision transformer using shifted windows.
\newblock In \emph{Proceedings of the IEEE/CVF international conference on computer vision}, pp.\  10012--10022, 2021.

\bibitem[Oord et~al.(2018)Oord, Li, and Vinyals]{oord2018representation}
Oord, A. v.~d., Li, Y., and Vinyals, O.
\newblock Representation learning with contrastive predictive coding.
\newblock \emph{arXiv preprint arXiv:1807.03748}, 2018.

\bibitem[Ouyang et~al.(2025)Ouyang, Hu, Zhang, Wang, and Wang]{ouyang2025projection}
Ouyang, Z., Hu, K., Zhang, Q., Wang, Y., and Wang, Y.
\newblock Projection head is secretly an information bottleneck.
\newblock \emph{arXiv preprint arXiv:2503.00507}, 2025.

\bibitem[Pedregosa et~al.(2011)Pedregosa, Varoquaux, Gramfort, Michel, Thirion, Grisel, Blondel, Prettenhofer, Weiss, Dubourg, et~al.]{pedregosa2011scikit}
Pedregosa, F., Varoquaux, G., Gramfort, A., Michel, V., Thirion, B., Grisel, O., Blondel, M., Prettenhofer, P., Weiss, R., Dubourg, V., et~al.
\newblock Scikit-learn: Machine learning in python.
\newblock \emph{the Journal of machine Learning research}, 12:\penalty0 2825--2830, 2011.

\bibitem[R{\'e}nyi(1961)]{renyi1961measures}
R{\'e}nyi, A.
\newblock On measures of entropy and information.
\newblock In \emph{Proceedings of the fourth Berkeley symposium on mathematical statistics and probability, volume 1: contributions to the theory of statistics}, volume~4, pp.\  547--562. University of California Press, 1961.

\bibitem[Roy \& Vetterli(2007)Roy and Vetterli]{roy2007effective}
Roy, O. and Vetterli, M.
\newblock The effective rank: A measure of effective dimensionality.
\newblock In \emph{2007 15th European signal processing conference}, pp.\  606--610. IEEE, 2007.

\bibitem[Saunshi et~al.(2019)Saunshi, Plevrakis, Arora, Khodak, and Khandeparkar]{saunshi2019theoretical}
Saunshi, N., Plevrakis, O., Arora, S., Khodak, M., and Khandeparkar, H.
\newblock A theoretical analysis of contrastive unsupervised representation learning.
\newblock In \emph{International Conference on Machine Learning}, pp.\  5628--5637. PMLR, 2019.

\bibitem[Shwartz-Ziv et~al.(2023)Shwartz-Ziv, Balestriero, Kawaguchi, Rudner, and LeCun]{shwartz2023information}
Shwartz-Ziv, R., Balestriero, R., Kawaguchi, K., Rudner, T.~G., and LeCun, Y.
\newblock An information-theoretic perspective on variance-invariance-covariance regularization.
\newblock \emph{arXiv preprint arXiv:2303.00633}, 2023.

\bibitem[Simon et~al.(2023)Simon, Knutins, Ziyin, Geisz, Fetterman, and Albrecht]{simon2023stepwise}
Simon, J.~B., Knutins, M., Ziyin, L., Geisz, D., Fetterman, A.~J., and Albrecht, J.
\newblock On the stepwise nature of self-supervised learning.
\newblock In \emph{International Conference on Machine Learning}, pp.\  31852--31876. PMLR, 2023.

\bibitem[Srinath~Halvagal et~al.(2023)Srinath~Halvagal, Laborieux, and Zenke]{srinath2023implicit}
Srinath~Halvagal, M., Laborieux, A., and Zenke, F.
\newblock Implicit variance regularization in non-contrastive ssl.
\newblock \emph{Advances in Neural Information Processing Systems}, 36:\penalty0 63409--63436, 2023.

\bibitem[Tan et~al.(2023)Tan, Yang, Huang, Yuan, and Zhang]{tan2023information}
Tan, Z., Yang, J., Huang, W., Yuan, Y., and Zhang, Y.
\newblock Information flow in self-supervised learning.
\newblock \emph{arXiv preprint arXiv:2309.17281}, 2023.

\bibitem[Thilak et~al.(2023)Thilak, Huang, Saremi, Dinh, Goh, Nakkiran, Susskind, and Littwin]{thilak2023lidar}
Thilak, V., Huang, C., Saremi, O., Dinh, L., Goh, H., Nakkiran, P., Susskind, J.~M., and Littwin, E.
\newblock Lidar: Sensing linear probing performance in joint embedding ssl architectures.
\newblock \emph{arXiv preprint arXiv:2312.04000}, 2023.

\bibitem[Tian et~al.(2021)Tian, Chen, and Ganguli]{tian2021understanding}
Tian, Y., Chen, X., and Ganguli, S.
\newblock Understanding self-supervised learning dynamics without contrastive pairs.
\newblock In \emph{International Conference on Machine Learning}, pp.\  10268--10278. PMLR, 2021.

\bibitem[Tishby et~al.(2000)Tishby, Pereira, and Bialek]{tishby2000information}
Tishby, N., Pereira, F.~C., and Bialek, W.
\newblock The information bottleneck method.
\newblock \emph{arXiv preprint physics/0004057}, 2000.

\bibitem[Uelwer et~al.(2025)Uelwer, Robine, Wagner, H{\"o}ftmann, Upschulte, Konietzny, Behrendt, and Harmeling]{uelwer2025survey}
Uelwer, T., Robine, J., Wagner, S.~S., H{\"o}ftmann, M., Upschulte, E., Konietzny, S., Behrendt, M., and Harmeling, S.
\newblock A survey on self-supervised methods for visual representation learning.
\newblock \emph{Machine Learning}, 114\penalty0 (4):\penalty0 1--56, 2025.

\bibitem[Van~Horn et~al.(2018)Van~Horn, Mac~Aodha, Song, Cui, Sun, Shepard, Adam, Perona, and Belongie]{van2018inaturalist}
Van~Horn, G., Mac~Aodha, O., Song, Y., Cui, Y., Sun, C., Shepard, A., Adam, H., Perona, P., and Belongie, S.
\newblock The inaturalist species classification and detection dataset.
\newblock In \emph{Proceedings of the IEEE conference on computer vision and pattern recognition}, pp.\  8769--8778, 2018.

\bibitem[Von~Neumann(2018)]{von2018mathematical}
Von~Neumann, J.
\newblock \emph{Mathematical foundations of quantum mechanics: New edition}.
\newblock Princeton university press, 2018.

\bibitem[Wang \& Isola(2020)Wang and Isola]{wang2020understanding}
Wang, T. and Isola, P.
\newblock Understanding contrastive representation learning through alignment and uniformity on the hypersphere.
\newblock In \emph{International conference on machine learning}, pp.\  9929--9939. PMLR, 2020.

\bibitem[Weng et~al.(2023)Weng, Ni, Song, Luo, Anwer, Khan, Khan, and Huang]{weng2023modulate}
Weng, X., Ni, Y., Song, T., Luo, J., Anwer, R.~M., Khan, S., Khan, F.~S., and Huang, L.
\newblock Modulate your spectrum in self-supervised learning.
\newblock \emph{arXiv preprint arXiv:2305.16789}, 2023.

\bibitem[Weng et~al.(2025)Weng, An, Ma, Qi, Luo, Yang, Dong, and Huang]{weng2025clustering}
Weng, X., An, J., Ma, X., Qi, B., Luo, J., Yang, X., Dong, J.~S., and Huang, L.
\newblock Clustering properties of self-supervised learning.
\newblock \emph{arXiv preprint arXiv:2501.18452}, 2025.

\bibitem[Yang et~al.(2023)Yang, Shi, Wei, Liu, Zhao, Ke, Pfister, and Ni]{yang2023medmnist}
Yang, J., Shi, R., Wei, D., Liu, Z., Zhao, L., Ke, B., Pfister, H., and Ni, B.
\newblock Medmnist v2-a large-scale lightweight benchmark for 2d and 3d biomedical image classification.
\newblock \emph{Scientific Data}, 10\penalty0 (1):\penalty0 41, 2023.

\bibitem[Yao \& Miller(2015)Yao and Miller]{yao2015tiny}
Yao, L. and Miller, J.
\newblock Tiny imagenet classification with convolutional neural networks.
\newblock \emph{CS 231N}, 2\penalty0 (5):\penalty0 8, 2015.

\bibitem[Zbontar et~al.(2021)Zbontar, Jing, Misra, LeCun, and Deny]{zbontar2021barlow}
Zbontar, J., Jing, L., Misra, I., LeCun, Y., and Deny, S.
\newblock Barlow twins: Self-supervised learning via redundancy reduction.
\newblock In \emph{International conference on machine learning}, pp.\  12310--12320. PMLR, 2021.

\bibitem[Zhan et~al.(2020)Zhan, Xie, Liu, Ong, and Loy]{zhan2020online}
Zhan, X., Xie, J., Liu, Z., Ong, Y.-S., and Loy, C.~C.
\newblock Online deep clustering for unsupervised representation learning.
\newblock In \emph{Proceedings of the IEEE/CVF conference on computer vision and pattern recognition}, pp.\  6688--6697, 2020.

\bibitem[Zhang et~al.(2023)Zhang, Tan, Yang, Huang, and Yuan]{zhang2023matrix}
Zhang, Y., Tan, Z., Yang, J., Huang, W., and Yuan, Y.
\newblock Matrix information theory for self-supervised learning.
\newblock \emph{arXiv preprint arXiv:2305.17326}, 2023.

\bibitem[Zheng et~al.(2021)Zheng, You, Wang, Qian, Zhang, Wang, and Xu]{zheng2021ressl}
Zheng, M., You, S., Wang, F., Qian, C., Zhang, C., Wang, X., and Xu, C.
\newblock Ressl: Relational self-supervised learning with weak augmentation.
\newblock \emph{Advances in Neural Information Processing Systems}, 34:\penalty0 2543--2555, 2021.

\end{thebibliography}
\bibliographystyle{icml2026}

%%%%%%%%%%%%%%%%%%%%%%%%%%%%%%%%%%%%%%%%%%%%%%%%%%%%%%%%%%%%

\appendix
\newpage

\section{Appendix Experimental Details}
\label{sec:appendex_experiments}

\subsection{Codebase}
\label{sec:codebase}

We use the solo-learn codebase \cite{da2022solo} and the codebase associated with the Resa algorithm  \cite{weng2025clustering}. Code for this paper will be released upon acceptance. The licenses for these repositories are MIT and Apache-2.0 respectively.

\subsection{Datasets}
\label{sec:datasets}
We show explicit details of all datasets used in this paper in Table \ref{tab:dataset_configuration}. The datasets were chosen to achieve as much diversity across a wide variety of data settings. This includes medical and natural image datasets, datasets of varying sizes, datasets of varying class complexity, and datasets with varying class imbalances. 

%\afterpage{
\begin{table}[h!]
\centering
\scalebox{.5}{
\begin{tabularx}{\textwidth}{ll|X|r}
Dataset & Abbreviation \& Link & Description & \# of classes \\
\hline
CIFAR-100 \cite{krizhevsky2009learning} & \href{https://www.cs.toronto.edu/~kriz/cifar.html}{cifar100} & 100 classes of 32x32 color images, including animals, vehicles, and various objects commonly found in the world. & 100 \\
\hline
CIFAR-10 \cite{krizhevsky2009learning} & \href{https://www.cs.toronto.edu/~kriz/cifar.html}{cifar10} & 10 classes of 32x32 color images featuring everyday objects and scenes such as airplanes, cars, and animals. & 10 \\
\hline
TinyImageNet200 \cite{yao2015tiny} & \href{http://tiny-imagenet.herokuapp.com/}{tinyimagenet200} & 200 classes of 64x64 images, a smaller version of the ImageNet dataset, used for object recognition and classification tasks. & 200 \\
\hline
BloodMNIST \cite{yang2023medmnist} & \href{https://medmnist.com/}{blood} & 8 classes of 28x28 images, designed for classification of diseases in red blood cells. & 8 \\\hline
OrganSMNIST \cite{yang2023medmnist} & \href{https://medmnist.com/}{organs} & 11 classes of 28x28 images, designed for classifying various types of liver tumor problems. & 11 \\\hline
OrganCMNIST \cite{yang2023medmnist} & \href{https://medmnist.com/}{organc} & 11 classes of 28x28 images, designed for classifying various types of liver tumor problems. & 11 \\\hline
OrganAMNIST \cite{yang2023medmnist} & \href{https://medmnist.com/}{organa} & 11 classes of 28x28 images, designed for classifying various types of liver tumor problems. & 11 \\\hline
STL10 \cite{yang2023medmnist} & \href{https://cs.stanford.edu/~acoates/stl10/}{stl10} & 10 classes of 96x96 images, designed for classifying various types of images. & 10 \\\hline
Cinic-10 \cite{darlow2018cinic} & \href{https://github.com/BayesWatch/cinic-10}{cinic10} & 10 classes of 96x96 images, designed for developing unsupervised feature learning, deep learning, and self-taught learning algorithms. & 10 \\
\hline
iNaturalist 2021 \cite{van2018inaturalist} & \href{https://www.kaggle.com/c/inaturalist-2021}{inat21} & Large-scale dataset with over 10,000 species, collected from photographs of plants and animals in their natural environments for fine-grained classification. & 10,000 \\
\hline
ImageNet \cite{deng2009imagenet} & \href{http://www.image-net.org/}{imagenet} & Large dataset with over 1,000 classes, used for image classification and object detection, containing millions of images across a wide variety of categories. & 1,000 \\
\hline

\end{tabularx}}
\caption{Overview of the datasets used in this paper.}
\label{tab:dataset_configuration}
\end{table}
\clearpage

\subsection{Pseudo-Code of AdaDim}
\label{sec:pseudocode}
\setminted{
    fontsize=\small,
    baselinestretch=0.8,
    breaklines=true,
    tabsize=2,
    framesep=1mm % If using frames
}
\begin{algorithm}[htp]
\caption{Pytorch style pseudocode for AdaDim}

\tiny
\begin{verbatim}

# E, Em: encoder, momentum encoder (optional)
# G, Gm: projector mlp, momentum projector mlp (optional)
# T1, T2: augmentation1, augmentation2
# temp: temperature = 0.1
# gama: gamma = 1e-4
# b: Batch size = 256
# b_i: Batch index in a specific epoch
# f_loss = feature decorrelation loss 
# s_loss = sample_uniformity loss
# alpha_renyi_loss with parameter of 2
# get_rank function computes effective rank
# alpha_epoch: epoch where the alpha parameter is computed
# current_epoch: Epoch of Current Training
import torch.nn.functional as F
############################
# Loss without momentum encoder
############################
rank_list = []
for x in loader: # load a minibatch x with b samples
    x1, x2 = T1(x), T2(x) # two augmentation views
    r1, r2 = E(x1), E(x2) # representation space
    z1, z2 = G(r1), G(r2) # embeddings 

    if current_epoch % alpha_epoch == 0 and b_i <=10:
        Z = torch.cat((z1,z2))
        min_dim = min(Z.shape[0],Z.shape[1])
        rank = get_rank(Z)
        rank_list.append(rank
    alpha = mean(rank_list) / min_dim
    beta = alpha * gamma

    f = f_loss(z1,z2)
    z1_norm = F.normalize(z1)
    z2_norm = F.normalize(z2)
    s = s_loss(z1_norm,z2_norm,temperature)
    ssl_loss = (1-alpha) * f + alpha * s
    mut_loss = (alpha_renyi_loss(z1,r2) + alpha_renyi_loss(z2,r1)) / 2
        
    return (1-beta) * ssl_loss  - beta * mut_loss

############################
# Loss with Momentum encoder
############################
rank_list = []
for x in loader: # load a minibatch x with b samples
    x1, x2 = T1(x), T2(x) # two augmentation views
    r1, r2 = E(x1), E(x2) # representation space
    z1, z2 = G(h1), G(h2) # embeddings 

    with torch.no_grad():
        update_momentum_params(0.996 -> 1) # exponential moving average
        r1m, r2m = Em(x1), Em(x2) # momentum encodings
        z1m, z2m = Gm(r1m), Gm(r2m) # momentum embeddings 

    if current_epoch % alpha_epoch == 0  and b_i <=10:
        Z = torch.cat((z1,z2m))
        min_dim = min(Z.shape[0],Z.shape[1])
        rank = get_rank(Z)
        rank_list.append(rank)
    alpha = mean(rank_list) / min_dim
    beta = alpha * gamma

    f = (f_loss(z1,z2m) + f_loss(z1m,z2)) / 2
    z1_norm = F.normalize(z1)
    z2_norm = F.normalize(z2)
    z1m_norm = F.normalize(z1m)
    z2m_norm = F.normalize(z2m)
    s = (s_loss(z1_norm,z2m_norm,temperature) + s_loss(z1m_norm,z2_norm,temperature)) / 2 
    ssl_loss = (1-alpha) * f + alpha * s
    mut_loss = (alpha_renyi_loss(z1,r2) + alpha_renyi_loss(z2,r1))
    mut_loss_momentum = (alpha_renyi_loss(z1m,r2m) + alpha_renyi_loss(z2m,r1m))
    total_mut_loss = (mut_loss + mut_loss_momentum) / 4
        
    return (1-beta) * ssl_loss  - beta * total_mut_loss

\end{verbatim}

\end{algorithm}

We note that AdaDim can be applied with an additional momentum encoder that is used in a variety of works \cite{chen2020improved,weng2025clustering}. In this setting, the momentum encoder is a separate network that is effectively a copy of the main backbone network. The difference is that a  momentum encoder is not updated with backpropagation, but is instead updated as an exponential moving average of the weights of the main backbone network. During training, the encoder produces its own representation matrices that have previously been shown to aid in representational stability when integrated with other SSL methods. We can then perform the same AdaDim loss with the additional momentum encoding representations and embeddings as shown in the pseudocode above. Note that our projector has an output dimension of 512 and the final layer is passed through a 1D batch normalization step.

\subsection{Method Training Details}
\label{sec:method_details}
\subsubsection{Baseline Setting}

We show the basic parameters for the baseline setting in Table \ref{tab:AdaDim_baseline} and the associated dataset specific parameters in Table \ref{tab:AdaDim_dataset}. This setting uses the basic joint embedding architecture setup without any additional architectural techniques such as a momentum encoder or predictor head. In this setting, we also use the LARS optimizer, a 256 batch size, 1e-4 weight decay, 10 epochs of warmup, a $\gamma=1e-4$, a temperature of 0.1, projector output of 2048, and an $\alpha$ update every 20 epochs. The augmentation scheme is the asymmetric augmentation scheme of \cite{da2022solo} described in Table~\ref{tab:augmentations}. 

For evaluation of our method, we use the online linear evaluation setting of \cite{da2022solo} where the classifier is trained alongside the backbone and projector during SSL pre-training. Representations are fed to a linear classifier while keeping the gradient of the classifier’s cross entropy loss from flowing through the backbone. The linear classifier has its own separate learning rate of 0.1 that follows a cosine annealing schedule based on the number of epochs of SSL training. The performance of the online classifier correlates well with the offline setting, making it a reliable proxy as shown in \cite{garrido2022duality,chen2020simple,da2022solo}. 

\subsubsection{Expanded Setting}

In the expanded setting, we use the same hyperparameters described in \cite{weng2025clustering} who also borrow the parameters of \cite{weng2023modulate}. These parameters are used when we integrate a momentum encoder or predictor head on top of the standard joint embedding architecture that we use in the baseline setting. The basic parameters for this setting are described in Table~\ref{tab:AdaDim_baseline} with dataset specific alterations described in Table~\ref{tab:AdaDim_dataset}. These parameters include a SGD optimizer, batch size of 256, base learning rate of 0.4, cosine scheduler, weight decay of 1e-4, 2 warmup epochs, $\gamma=1e-4$, temperature of 0.1, 512 output projector size and 20 epochs between every $\alpha$ update. The attached momentum encoder also has a momentum encoding update parameter of 0.996. This setting also uses the augmentation scheme of \cite{zheng2021ressl} described in Table~\ref{tab:augmentations}.

Another difference in this setting is that evaluation is done in an offline manner where a linear layer is appended to the frozen SSL encoder and fine-tuned with specific hyperparameters. Specifically, we use the linear fine-tuning approach of \cite{weng2025clustering} and \cite{weng2023modulate} described in Table~\ref{tab:AdaDim_momentum_linear}. For small scale experiments (i.e. datasets that are notImageNet), the linear layer is fine-tuned for 500 epochs, with an Adam optimizer, a learning rate that drops from 1e-2 to 1e-6, and a weight decay of 5e-6. Training is performed in a stochastic manner where subsets of the training set are used at each epoch. This setting allows evaluation to take place in a few minutes and is suitable for analyzing smaller datasets. For ImageNet and ImageNet-100 experiments, an SGD optimizer is used with a momentum parameter of 0.9, 100 epochs of training, a base learning rate and weight decay specified in Table~\ref{tab:AdaDim_dataset}, and step down  scheduler by a factor of 10 at epochs 60 and 80.

\subsubsection{Comparison Methods}

All essential hyperparameters for comparisons with  state of the art methods are shown in Table \ref{tab:comparison_methods}. Note that these parameters are used to train comparisons for the ablation study of Table~\ref{tab:parameter_var} and the diverse data study of Table~\ref{tab: dataset_wrapped}. For all other comparisons against past state of the art SSL methods, we use the benchmark tables provided in \cite{da2022solo} for comparisons with Cifar-10, Cifar-100, and ImageNet-100. For comparisons with ImageNet, we copy the tables provided in \cite{weng2025clustering} that is taken from the results reported in the original paper for each method. In the case of method specific hyperparameters, we use the parameters described in the solo-learn codebase as much as possible. 

In Table~\ref{tab:parameter_var}, we also compare with the regularization technique of \cite{ouyang2025projection}.  This involves taking the matrices $R$ and $Z$ and computing the mutual information estimate based on the $\alpha$-Renyi approximation discussed in Section \ref{sec:metrics}. This regularization term is scaled by a $\lambda$ parameter that is set to 1e-4 for all experiments. This specific choice of $\lambda$ is based on the value that performed best in  \cite{ouyang2025projection}.

\begin{table}[]
\scalebox{0.63}{
\begin{tabular}{ccccccccccccc}
\hline
\multicolumn{13}{c}{AdaDim Baseline Parameters}                                                                                          \\ \hline
Method & Setting   & M & P & Optimizer & Batch Size & Base LR & Weight Decay & Warmup & $\gamma$ & Temp & \multicolumn{1}{l}{Alpha Update} & Proj Output \\ \hline
AdaDim & Baseline & \xmark & \xmark & LARS      & 256        & 0.4     & 1e-4         & 10     & 1e-4  & 0.1  & 20  & 2048                             \\
AdaDim & Expanded       & \checkmark & \checkmark or \xmark & SGD       & 256        & 0.4     & 1e-4         & 2      & 1e-4  & 0.1  & 20   & 512                          \\ \hline
\end{tabular}}
\caption{This table shows the baseline parameters for most experiments. M refers to using a momentum encoder. P refers to using an additional prediction head.}
\label{tab:AdaDim_baseline}
\end{table}

\begin{table}[]
\scalebox{0.65}{
\begin{tabular}{cccccccccc}
\hline
\multicolumn{10}{c}{AdaDim Expanded Hyperparameter Specific Details}                                                                                                                                                              \\ \hline
Method & Setting & M & P & \begin{tabular}[c]{@{}c@{}}Pretraining\\ Momentum\end{tabular} & Linear Optimizer & Linear Base LR & Linear Weight Decay & Linear Scheduler & \begin{tabular}[c]{@{}c@{}}Linear\\ Epochs\end{tabular} \\ \hline
AdaDim & Expanded     & \checkmark & \checkmark or \xmark & .996                                                           & Adam             & 0.01           & 5e-6                & Exponential      & 500                                                     \\ \hline
\end{tabular}}
\caption{This table shows the momentum parameter used during SSL pre-training within a momentum encoder and the details of the offline linear evaluation setting for the momentum encoding experiments. M refers to using a momentum encoder. P refers to using an additional prediction head.}
\label{tab:AdaDim_momentum_linear}
\end{table}
\begin{table}[h!]
\scalebox{0.55}{
\begin{tabular}{cccccccccccccc}
\hline
\multicolumn{13}{c}{AdaDim Dataset Specific Training}                                                                                                       \\ \hline
Method & Setting  & Dataset         & M & P & Tuned Gamma & Epochs & Model     & Batch Size & \begin{tabular}[c] {@{}c@{}}Augmentation\\ Scheme\end{tabular} &  Base Linear LR & Pretrain LR & WD\\ \hline
AdaDim & Baseline & Cifar100        & \xmark & \xmark & 1e-4        & 400    & ResNet-50 & 256        & Asymmetric  & 0.1 & 0.4 & 1e-4                                                 \\
AdaDim & Expanded & Cifar100        & \checkmark & \xmark & 1e-4        & 400    & ResNet-50 & 256        & ReSSL  & 0.01 & 0.4 & 1e-4                                                 \\
AdaDim & Baseline & TinyImageNet200 & \xmark & \xmark & 1e-4        & 400    & ResNet-50 & 256        & Asymmetric  & 0.1 & 0.4 & 1e-4                                                 \\
AdaDim & Expanded & TinyImageNet200        & \checkmark & \xmark & 1e-4        & 400    & ResNet-50 & 256        & ReSSL  & 0.01 & 0.4 & 1e-4                                                 \\
AdaDim & Baseline & Cinic10         & \xmark & \xmark & 1e-2        & 400    & ResNet-50 & 256        & Asymmetric  & 0.1   & 0.4 & 1e-4                                               \\
AdaDim & Expanded & Cinic10        & \checkmark & \xmark & 1e-4        & 400    & ResNet-50 & 256        & ReSSL  & 0.01 & 0.4 & 1e-4                                                 \\
AdaDim & Baseline & STL10           & \xmark & \xmark & 1e-2        & 400    & ResNet-50 & 256        & Asymmetric  & 0.1  & 0.4 & 1e-4                                                \\
AdaDim & Expanded & STL10        & \checkmark & \xmark & 1e-4        & 400    & ResNet-50 & 256        & ReSSL  & 0.01 & 0.4 & 1e-4                                                 \\
AdaDim & Baseline & BloodMNIST      & \xmark & \xmark & 1e-4        & 400    & ResNet-50 & 256        & Asymmetric  & 0.1  & 0.4 & 1e-4                                                \\
AdaDim & Expanded & BloodMNIST        & \checkmark & \xmark & 1e-4        & 400    & ResNet-50 & 256        & ReSSL  & 0.01 & 0.4 & 1e-4                                                 \\
AdaDim & Baseline & OrganSMNIST     & \xmark & \xmark &1e-4        & 400    & ResNet-50 & 256        & Asymmetric  & 0.1    & 0.4 & 1e-4                                              \\
AdaDim & Expanded & OrganSMNIST        & \checkmark & \xmark & 1e-4        & 400    & ResNet-50 & 256        & ReSSL  & 0.01 & 0.4 & 1e-4                                                 \\
AdaDim & Baseline & OrganCMNIST     & \xmark & \xmark & 1e-2        & 400    & ResNet-50 & 256        & Asymmetric  & 0.1  & 0.4 & 1e-4                                                \\
AdaDim & Expanded & OrganCMNIST        & \checkmark & \xmark & 1e-4        & 400    & ResNet-50 & 256        & ReSSL  & 0.01 & 0.4 & 1e-4                                                 \\
AdaDim & Baseline & OrganAMNIST     & \xmark & \xmark & 1e-1        & 400    & ResNet-50 & 256        & Asymmetric    & 0.1 & 0.4 & 1e-4                                               \\
AdaDim & Expanded & OrganAMNIST        & \checkmark & \xmark & 1e-4        & 400    & ResNet-50 & 256        & ReSSL  & 0.01 & 0.4 & 1e-4                                                 \\
AdaDim & Baseline & Cifar100        & \xmark & \xmark & 1e-4        & 1000   & ResNet-18 & 256        & Asymmetric   & 0.1  & 0.4 & 1e-4                                               \\
AdaDim & Baseline & Cifar10         & \xmark & \xmark & 1e-4        & 1000   & ResNet-18 & 256        & Asymmetric    & 0.1 & 0.4 & 1e-4                                               \\
AdaDim & Baseline & ImageNet100     & \xmark & \xmark & 1e-1       & 400    & ResNet-18 & 256        & Asymmetric  & 0.1  & 0.4 & 1e-4                                                \\
AdaDim & Expanded & Cifar100        & \checkmark & \xmark & 1e-4        & 1000   & ResNet-18 & 256        & ReSSL    & 0.01  & 0.4 & 1e-4                                                   \\
AdaDim & Expanded & Cifar10          & \checkmark & \xmark & 1e-4        & 1000   & ResNet-18 & 256        & ReSSL    & 0.01   & 0.4 & 1e-4                                                   \\
AdaDim & Expanded & ImageNet100      & \checkmark & \xmark & 1e-4        & 400    & ResNet-18 & 128        & ReSSL    & 5      & 0.5 & 2.5e-5                                              \\
AdaDim & Expanded & ImageNet         & \checkmark & \xmark & 1e-4        & 100    & ResNet50  & 256        & ReSSL    & 2    & 0.5 & 1e-5                                                \\
AdaDim & Expanded & ImageNet        & \checkmark & \xmark & 1e-4        & 100    & ResNet50  & 256        & ReSSL  & 1     & 0.5 & 1e-5        
\\
AdaDim & Expanded & ImageNet        & \checkmark & \xmark & 1e-4        & 200    & ResNet50  & 256        & ReSSL  & 1     & 0.5 & 1e-5    
\\\hline
\end{tabular}}
\caption{This table shows the details of specific choices made on a per dataset basis for the tables generated in the main paper. M refers to using a momentum encoder. P refers to using an additional prediction head.}
\label{tab:AdaDim_dataset}
\end{table}
\begin{table}[]
\scalebox{0.55}{
\begin{tabular}{cccccccccc}
\hline
\multicolumn{10}{c}{Comparison Method Parameters}                                                                                                                                                                                                                                                                                                                        \\ \hline
Method         & Setting  & \begin{tabular}[c]{@{}c@{}}Specific\\ Parameters\end{tabular} & Model     & Epochs & Optimizer & \begin{tabular}[c]{@{}c@{}}Learning\\ Rate\end{tabular} & \begin{tabular}[c]{@{}c@{}}Batch\\ Size\end{tabular} & \begin{tabular}[c]{@{}c@{}}Augmentation\\ Scheme\end{tabular} & \begin{tabular}[c]{@{}c@{}}Projector\\ Output\end{tabular} \\ \hline
SimCLR         & Baseline & temp = 0.1                                                    & ResNet-50 & 400    & LARS      & 0.4                                                     & 256                                                  & Symmetric                                                     & 128                                                        \\
VICReg         & Baseline & Params= 25,25,1                                               & ResNet-50 & 400    & LARS      & 0.4                                                     & 256                                                  & Asymmetric                                                    & 2048                                                       \\
Moco v2        & Baseline & temp = 0.2, momentum = {[}0.9,0.99{]}, queue = 32768          & ResNet-50 & 400    & SGD       & 0.3                                                     & 256                                                  & weak\_symmetric                                               & 256                                                        \\
BYOL           & Baseline & momentum = 1.0, base  = 0.99                                  & ResNet-50 & 400    & LARS      & 1.0                                                     & 256                                                  & asymmetric                                                    & 256                                                        \\
Barlow Twins   & Baseline & scale\_loss = 0.1                                             & ResNet-50 & 400    & LARS      & 0.3                                                     & 256                                                  & asymmetric                                                    & 2048                                                       \\
NNCLR          & Baseline & queue = 65536, temp = 0.2                                     & ResNet-50 & 400    & LARS      & 0.4                                                     & 256                                                  & asymmetric                                                    & 256                                                        \\
SimSiam        & Baseline & temp = 0.2                                                    & ResNet-50 & 400    & LARS      & 0.5                                                     & 256                                                  & weak\_symmetric                                               & 512                                                        \\
DeepCluster v2 & Baseline & prototypes = {[}3000, 3000, 3000{]}                           & ResNet-50 & 400    & LARS      & 0.6                                                     & 256                                                  & symmetric                                                     & 128                                                        \\
Moco v3        & Baseline & momentum = {[}0.9, 0.99{]}                                    & ResNet-50 & 400    & LARS      & 0.3                                                     & 256                                                  & asymmetric                                                    & 256                                                        \\ \hline
\end{tabular}}
\caption{This table shows the training details of the models we trained directly for the diverse dataset study in the main paper. The same hyperparameters were maintained across all datasets.}
\label{tab:comparison_methods}
\end{table}

\begin{table}[h!]
\begin{tabular}{ccccc}
\hline
\multicolumn{5}{c}{Augmentation Details}                                                                  \\ \hline
                  & \multicolumn{2}{c|}{Asymmetric Augmentation} & \multicolumn{2}{c}{ReSSL Augmentation} \\ \hline
Augmentation Type & View 1     & \multicolumn{1}{c|}{View 2}     & View 1             & View 2            \\ \hline
Crop              & 1          & 1                               & 1                  & 1                 \\
Color Jitter      & 0.8        & 0.8                             & 0.0                & 0.8               \\
Contrast          & 0.4        & 0.4                             & 0.0                & 0.4               \\
Brightness        & 0.4        & 0.4                             & 0.0                & 0.4               \\
Saturation        & 0.2        & 0.2                             & 0.0                & 0.2               \\
Hue               & 0.1        & 0.1                             & 0.0                & 0.1               \\
Grayscale         & 0.2        & 0.2                             & 0.0                & 0.2               \\
Gaussian Blur     & 1.0        & 0.1                             & 0.0                & 0.0               \\
Solarization      & 0.0        & 0.2                             & 0.0                & 0.2               \\
Equalization      & 0.0        & 0.0                             & 0.0                & 0.0               \\
Horizontal Flip   & 0.5        & 0.5                             & 0.5                & 0.5               \\ \hline
\end{tabular}
\caption{In this table, we detail the augmentations applied to each view (View 1 and 2) during pre-training of our algorithm. We divide the augmentations based on the type of scheme we are using. Every value in the table represents the probability of a specific augmentation type being applied.}
\label{tab:augmentations}
\end{table}

\subsection{Sample Uniformity and Feature Decorrelation Losses}
\label{sec:sim_vicreg}

In this section, we go into more depth regarding the $L_{SU}$ and $L_{FD}$ losses. Suppose there is an image $i$ drawn from a training pool $i \in I$. $i$ is passed into two random transformations $t(i) = x$ and $t^{'}(i) = x^{'}$ where $t$ and $t^{'}$ are drawn from the set of all random augmentations $T$. Both $x$ and $x^{'}$ are passed into an encoder network $e(\cdot)$. This results in the representations $e(x) = r$ and $e(x^{'})  = r^{'}$. These representations are then passed into a projection head $g(\cdot)$ that produces the embeddings $g(x) = z$ and $g(x^{'}) = z^{'}$. The collection of all representations and embeddings within a batch of $n$ samples can be represented by the $R$, $R^{'}$, $Z$, and $Z^{'}$ matrices. In this case, all matrices are composed of $n$ vectors with dimension $D$. This can be written as $R = [r_1, r_2, ..., r_n]$, $R^{'} = [r^{'}_1, r^{'}_2, ..., r^{'}_n]$, $Z^{'} = [z^{'}_1, z^{'}_2, ..., z^{'}_n]$, and $Z = [z_1, z_2, ..., z_n]$. The $L_{FD}$\cite{bardes2021vicreg} and $L_{SU}$ \cite{oord2018representation,chen2020simple} losses can be computed. In this case, $L_{FD}$ corresponds to a feature decorrelation loss that is better at promoting higher $H(R)$ while $L_{SU}$  corresponds to a sample uniformity loss better at promoting lower $I(R;Z)$ at the end of training.  The $L_{SU}$  loss is written as: $ L_{SU} = -\sum_{i\in I} log\frac{exp(sim(z_i,z^{'}_i)/\tau)}{\sum_{k=1}^{2N} \mathbbm{1}[k \neq i] exp(sim(z_i,z_k)/\tau)}$ where $sim$ refers to the cosine similarity, $\tau$ represents a temperature parameter, and the summation in the denominator takes place over all samples from both transformations. The $L_{FD}$ loss is written as: $L_{FD} = \lambda s(Z,Z^{'}) + \mu [v(Z) + v(Z^{'})] + \nu[c(Z) + c(Z^{'})]$. The invariance term is $s(Z,Z^{'}) = \frac{1}{N}\sum_{i=1}^{N}||z_i-z^{'}_i||^{2}_2$. The covariance term is $c(Z) = \frac{1}{F}\sum_{i\neq j}[C(Z)]_{i,j}^2$ where $C(Z)$ is the covariance matrix of $Z$. The variance term is $v(Z) = \frac{1}{F}\sum_{j=1}^{F} max(0,\gamma-S(z^{j},\epsilon))$ where $S(x,\epsilon)$ is the regularized standard deviation, $z^{j}$ represents the vector of each value at dimension $j$, and $\gamma$ is a target value set to 1 for all experiments. For both $L_{SU}$ and $L_{FD}$, we use the conventions of the original papers which includes $\tau = 0.1$, $\lambda = \mu = 25$, and $\nu = 1$. 

Note that other losses for feature decorrelation and sample uniformity are also possible. However, we found that using these terms from \cite{oord2018representation} and \cite{bardes2021vicreg} resulted in a loss setup that was efficient to compute due to reusing many of the same terms across both calculations.

\subsection{Metric Analysis Details}
\label{sec:metrics}

One possible mathematical description for the dimensionality of a representation space $H(R)$ is the von Neumann entropy of eigenvalues \cite{von2018mathematical,kim2023vne} which takes the form $H(R) = -\sum_{i} \lambda_i log(\lambda_i)$ where each $\lambda_i$ represents an eigenvalue of $R$. To increase $H(R)$ in this formula, we can either increase the total number of non-zero eigenvalues or maintain the same number of eigenvalues, but make the eigenvalues more similar in value to each other (higher uniformity, lower variance). Increasing the total number of eigenvalues corresponds to feature decorrelation in which an SSL algorithm discovers a larger number of total dimensions along which $R$ can vary. Decreasing the variance of eigenvalues within a fixed dimensional space corresponds to sample uniformity where representations spread more equally along all dimensions.

Throughout the paper, the dynamics between $H(R)$ and $I(R;Z)$ is discussed. However, this analysis requires a variety of metrics that were not fully detailed in the main paper. For our analytical experiments, the test set of interest is passed into the trained SSL model and its associated projection head. This results in a matrix for the representation space $R$ and a corresponding matrix for the embedding space $Z$. These matrices are of size:  number of test set samples $\times$ 2048. These matrices are then used to compute the metrics used for analysis in the main paper. $I(R;Z)$ is computed using the $\alpha$-Renyi matrix mutual information approximation discussed in \cite{tan2023information}. To calculate this quantity, assume that normalized matrices A and B are both $R^{nxn}$. The entropy of matrix A can be represented as $H_{\alpha}(A) = \frac{1}{1-\alpha}log[tr((\frac{A}{n})^{\alpha})]$ where $\alpha$ =2 for all experiments. This formulation results in a matrix mutual information estimator of the form $I(A;B) = H_{\alpha}(A) +  H_{\alpha}(B) - H_{\alpha}(A\odot B)$ where $\odot$ is the hadamard product. This formulation only works for positive semi definite matrices so during our experiments the approximation of \cite{tan2023information} is followed where the normalized covariance matrices $RR^T$ and $ZZ^{T}$ are used as inputs to calculate $I(R;Z)$. 

\begin{figure}[h!]
    \centering
    \includegraphics[width=\linewidth]{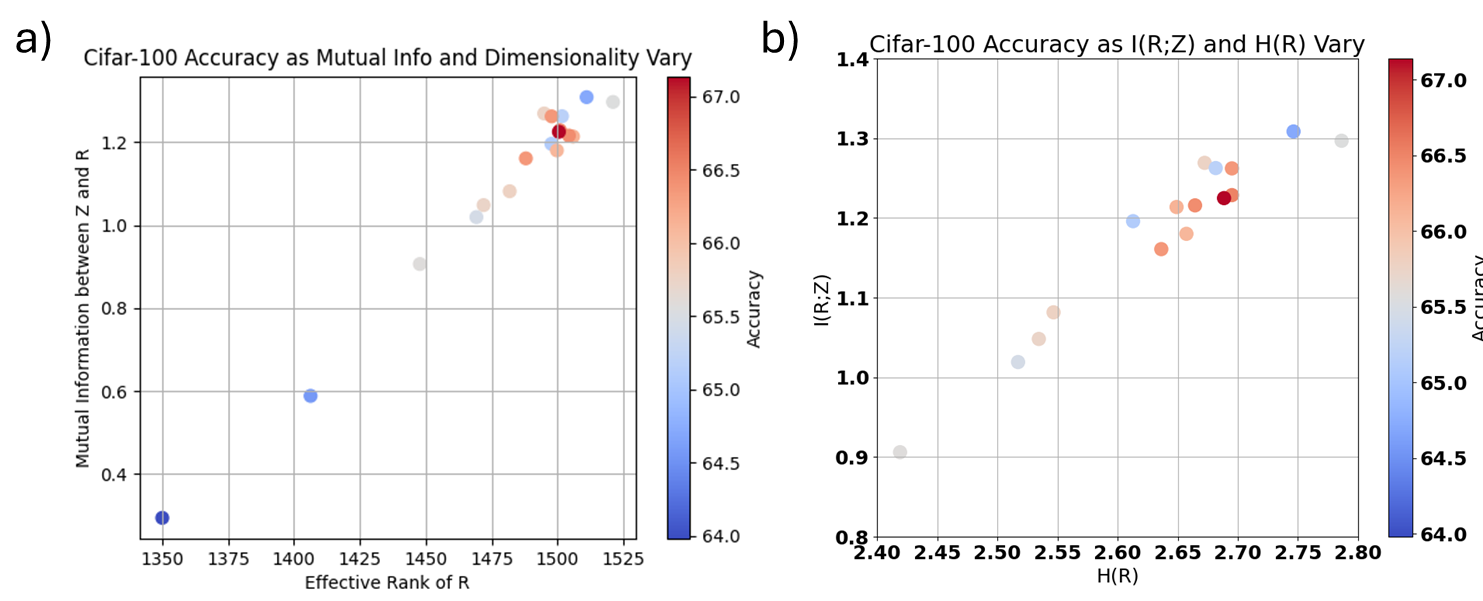}
    \caption{We show versions of the same opening Figure with $H(R)$ computed with a) the effective rank and b) an $\alpha$-Renyi matrix approximator.}
    \label{fig:trade-off_compare}
\end{figure}

Note that there are a variety of ways to approximate $H(R)$. In this paper, both $H_{\alpha}(R)$ and the effective rank\cite{roy2007effective} are used at different points. The main reason for this choice is that the effective rank is normalized with respect to the eigenvalues of the current distribution. This means that the lowest possible value is 0 and the highest possible value is minimum dimension of the matrix of interest. The advantage of the $\alpha$-Renyi approximator is that the scale of the values will more closely match the values used to calculate $I(R;Z)$. However, both metrics result in the same balancing behavior between $H(R)$ and $I(R;Z)$ and are correlated with each other. This correlation is observed in Figure \ref{fig:trade-off_compare}. In general, any computation of $H(R)$ can be thought of as an approximation of the dimensionality of the representation space. This is because higher dimensionality has been characterized in terms of eigenvalue distributions across a variety of works \cite{garrido2022duality,thilak2023lidar,agrawal2022alpha,jing2021understanding}. These metrics follow this trend as they are based on measuring how closely the eigenvalue spectrum of a given matrix approaches a uniform distribution. For example, another possible entropy estimator is discussed in \cite{zhang2023matrix}. This work states that for a positive semi definite (PSD) matrix $A$, matrix entropy (ME) can be defined as $ME(A) = -tr(A log(A)) + tr(A) = -\sum_i\lambda_ilog(\lambda_i) + \sum_i\lambda_i$. The first term will increase with the sample uniformity of the representation space i.e. as the eigenvalues become more uniformly distributed. The second term will increase with more and larger eigenvalues i.e. as the features of the space become more decorrelated.

The uniformity metric \cite{wang2020understanding} is also used as part of our analysis. This metric acts as a measurement of how uniformly distributed the points of a representation space are on a hypersphere. It takes the form of the pairwise gaussian potential kernel  and can be expressed as $ log(\mathbb{E}_{(x,y) \sim p_{data}}[e^{-2||e(x)-e(y)||^2_{2}}])$. In general, greater uniformity indicates a more negative value when this metric is computed empirically. We use the implementation from the original \href{https://github.com/ssnl/align_uniform.git}{github} of \cite{wang2020understanding}.

\subsection{Compute Resources}
\label{sec:compute_resources}
Our resources included a personal PC with 8 Intel i7-6700K CPU Cores and 2 12 GB Nvidia GeForce GTX Titan X GPUs. We also used a lab work station server with 12 Intel i7-5930K CPU cores and 2 24GB Nvidia TITAN RTX 3090 GPUs. We also used a server with compute resources based on availability and priority queues. The vast majority of experiments use these resources and are found in the main paper or the appendix. However, there may be early exploratory experiments in the development of our method that were not included.

\begin{table}[]
\begin{tabular}{cccccc}
\hline
\multicolumn{6}{c}{Compute Analysis of 100 Epochs of Cifar-100 Training} \\ \hline
            & SimCLR     & VICReg     & AdaDim    & AdaDim + momentum  & Resa   \\ \hline
s/epoch     & 22.01      & 19.09      & 23.48     & 27.82    & 75.10            \\ \hline
\end{tabular}
\caption{This table shows the compute cost of 100 epochs of training of Cifar-100 with a ResNet-18 model. We compute the total time for the experiment and divide by the number of epochs to get the seconds/epoch metric (s/epoch). All experiments are performed in identical settings with the same number of CPU cores and the usage of an RTX 3090 GPU. For these experiments, the same data augmentation strategy was applied for ease of comparability.}
\label{tab:compute_experiment}
\end{table}

\subsection{Compute Discussion of our Method}
\label{sec:compute_method}

Our method involves computing the eigenspectrum at different points in the training process. In general, computing eigenvalues is an expensive operation with order $O(n^3)$. However, the number of calculations is limited through a few mechanisms specific to \texttt{AdaDim}. This includes the usage of the $E_{\alpha}$ parameter. This parameter dictates how many epochs must pass before the $\alpha$ parameter is re computed. In Figure \ref{fig:hyparameter_var_2}, performance improvements are maintained even when $E_{\alpha}$ is varied.  Additionally, for every $E_{\alpha}$, eigenvalues are computed for only 10 training batches. This design choice is based on the  empirical observation that most batches have a similar effective rank as training progresses. This limits the need to compute the eigenvalues across all batches in an epoch. The averaging across 10 batches is done to ensure additional stability regarding the $\alpha$ update. However, it may be possible to use even fewer batches in this computation.

To further analyze the compute cost of our method, we perform an experiment where we measure the amount of time it takes to train a ResNet-18 model for 100 epochs on Cifar-100 with the same compute resources across all methods of interest. This setup is similar to the compute analysis of other papers such as \cite{weng2025clustering}. The results of this experiment are shown in Table~\ref{tab:compute_experiment}. We observe that AdaDim does cost more than the baseline SimCLR or VICReg methods. This cost requirement increases further when adding an additional momentum encoder. However, the cost increase is not so drastic to make training with AdaDim a prohibitive endeavor. We observe minor increases in compute cost of 3 - 8 s/epoch above the baseline SimCLR approach, depending on whether we use the baseline AdaDim strategy or an additional momentum encoder network. Note that SimCLR, ViCReg, and the baseline version of AdaDim do not use any compute intensive techniques that are associated with other algorithms. In general, using these techniques causes a large increase in compute cost. To make this point clear, we perform the same experimental setup of Table 11 in the appendix with the most recent state of the art Resa algorithm that uses both a momentum queue and a clustering step with the sinkhorn-knopp algorithm. This results in a compute cost that is more than  double the next closest algorithm.

\subsection{Empirical Eigenvalue Analysis Details}
\label{sec:eig_analysis}

In Figure \ref{fig:2000_eig_simulation}, we perform a variety of analyses on the eigenvalue distribution of the output matrices of a ResNet-50 model. In part b), all eigenvalues are normalized before counting the number of eigenvalues above a threshold $\tau$ that we set to .01 for all experiments. This normalization is done by dividing all eigenvalues by the l-1 norm of the total eigenvalue spectrum. This is similar to the normalization done in the computation of the effective rank. In part c), the cumulative explained variance ratio metric is computed. To compute this metric, assume that there is a set of eigenvalues $\lambda = [\lambda_1, \lambda_2, ..., \lambda_N]$ where the eigenvalues are ordered from largest to smallest. Assume that there is a percentage $p$ of the largest eigenvalues. This results in the explained variance metric: $\frac{\sum_{i=1}^{\lfloor p*N \rfloor}\lambda_i}{\sum_{i=1}^{N}\lambda_i}$. This metric increases as the subset of eigenvalues that we sum over constitutes more of the overall variance of the data. However, it will decrease as the spread of this variance is distributed over eigenvalues outside of the percentage that the numerator is summed over.

\subsection{Random Parameter Ablation Study}

In Figure \ref{fig:random}, we show how accuracy varies as a function of $H(R)$ and $I(R;Z)$ for a variety of models with different hyperparameters. We also compute the correlation of these results with metrics of interest in Table~\ref{tab:corr}. We generate 15 models for 3 different methods on Cifar-100 and display the exact parameters for each of these methods in Table~\ref{tab:random_list}.

\begin{table}[]
\scalebox{0.7}{
\begin{tabular}{cccccccccc}
\hline
Method & Dataset   & Epochs & Parameters & Learning Rate & Temperature & Weight Decay & Effective Rank & Mutual Info & Accuracy \\ \hline
SimCLR & Cifar-100 & 100    & d=2048     & 0.6           & 0.05        & 10-6         & 673            & 0.073       & 47.15    \\
SimCLR & Cifar-100 & 100    & d=2048     & 0.6           & 0.07        & 10-6         & 682            & 0.071       & 48.84    \\
SimCLR & Cifar-100 & 100    & d=2048     & 0.6           & 0.1         & 10-6         & 684            & 0.079       & 49.43    \\
SimCLR & Cifar-100 & 100    & d=2048     & 0.6           & 0.2         & 10-6         & 646            & 0.075       & 52.28    \\
SimCLR & Cifar-100 & 100    & d=2048     & 0.6           & 0.3         & 10-6         & 594            & 0.076       & 54.39    \\
SimCLR & Cifar-100 & 100    & d=2048     & 0.6           & 0.4         & 10-6         & 566            & 0.068       & 51.99    \\
SimCLR & Cifar-100 & 100    & d=2048     & 0.5           & 0.05        & 10-6         & 658            & 0.064       & 48.65    \\
SimCLR & Cifar-100 & 100    & d=2048     & 0.5           & 0.07        & 10-6         & 652            & 0.056       & 47.36    \\
SimCLR & Cifar-100 & 100    & d=2048     & 0.5           & 0.1         & 10-6         & 670            & 0.055       & 54.56    \\
SimCLR & Cifar-100 & 100    & d=2048     & 0.5           & 0.15        & 10-6         & 666            & 0.074       & 54.03    \\
SimCLR & Cifar-100 & 100    & d=2048     & 0.5           & 0.2         & 10-6         & 637            & 0.078       & 54.98    \\
SimCLR & Cifar-100 & 100    & d=2048     & 0.5           & 0.3         & 10-6         & 587            & 0.078       & 54.15    \\
SimCLR & Cifar-100 & 100    & d=2048     & 0.5           & 0.4         & 10-6         & 556            & 0.07        & 53.1     \\
SimCLR & Cifar-100 & 100    & d=2048     & 0.5           & 0.15        & 10-7         & 655            & 0.076       & 54.86    \\
SimCLR & Cifar-100 & 100    & d=2048     & 0.5           & 0.15        & 10-5         & 666.67         & 0.076       & 53.59    \\
VICReg & Cifar-100 & 100    & nu = 0.3   & 0.3           & N/A         & 10-6         & 922            & 0.268       & 50.75    \\
VICReg & Cifar-100 & 100    & nu = 0.4   & 0.3           & N/A         & 10-6         & 914            & 0.265       & 50.83    \\
VICReg & Cifar-100 & 100    & nu = 0.5   & 0.3           & N/A         & 10-6         & 902            & 0.292       & 50.62    \\
VICReg & Cifar-100 & 100    & nu = 0.6   & 0.3           & N/A         & 10-6         & 892            & 0.263       & 52.29    \\
VICReg & Cifar-100 & 100    & nu = 0.7   & 0.3           & N/A         & 10-6         & 902            & 0.235       & 56.72    \\
VICReg & Cifar-100 & 100    & nu = 0.8   & 0.3           & N/A         & 10-6         & 903            & 0.24        & 56.27    \\
VICReg & Cifar-100 & 100    & nu = 0.9   & 0.3           & N/A         & 10-6         & 898            & 0.237       & 55.73    \\
VICReg & Cifar-100 & 100    & nu = 1.0   & 0.3           & N/A         & 10-6         & 883            & 0.244       & 52.37    \\
VICReg & Cifar-100 & 100    & nu = 1.1   & 0.3           & N/A         & 10-6         & 878            & 0.229       & 57.6     \\
VICReg & Cifar-100 & 100    & nu = 1.2   & 0.3           & N/A         & 10-6         & 877            & 0.232       & 52.28    \\
VICReg & Cifar-100 & 100    & nu = 1.3   & 0.3           & N/A         & 10-6         & 847            & 0.257       & 54.54    \\
VICReg & Cifar-100 & 100    & nu = 1.4   & 0.3           & N/A         & 10-6         & 868            & 0.212       & 55.09    \\
VICReg & Cifar-100 & 100    & nu = 1.5   & 0.3           & N/A         & 10-6         & 795            & 0.279       & 49.19    \\
VICReg & Cifar-100 & 100    & nu = 1.6   & 0.3           & N/A         & 10-6         & 867            & 0.209       & 54.48    \\
VICReg & Cifar-100 & 100    & nu = 1.7   & 0.3           & N/A         & 10-6         & 848            & 0.2107      & 58.69    \\
NNCLR  & Cifar-100 & 100    & d=2048     & 0.6           & 0.05        & 10-6         & 416            & 0.043       & 54.09    \\
NNCLR  & Cifar-100 & 100    & d=2048     & 0.6           & 0.07        & 10-6         & 417            & 0.038       & 54.27    \\
NNCLR  & Cifar-100 & 100    & d=2048     & 0.6           & 0.1         & 10-6         & 459            & 0.033       & 55.47    \\
NNCLR  & Cifar-100 & 100    & d=2048     & 0.6           & 0.2         & 10-6         & 493            & 0.058       & 55.9     \\
NNCLR  & Cifar-100 & 100    & d=2048     & 0.6           & 0.3         & 10-6         & 490            & 0.067       & 54.43    \\
NNCLR  & Cifar-100 & 100    & d=2048     & 0.6           & 0.4         & 10-6         & 519            & 0.067       & 55.07    \\
NNCLR  & Cifar-100 & 100    & d=2048     & 0.5           & 0.05        & 10-6         & 425            & 0.042       & 54.59    \\
NNCLR  & Cifar-100 & 100    & d=2048     & 0.5           & 0.07        & 10-6         & 439            & 0.036       & 55.16    \\
NNCLR  & Cifar-100 & 100    & d=2048     & 0.5           & 0.1         & 10-6         & 462            & 0.033       & 56.01    \\
NNCLR  & Cifar-100 & 100    & d=2048     & 0.5           & 0.15        & 10-6         & 474            & 0.05        & 56.09    \\
NNCLR  & Cifar-100 & 100    & d=2048     & 0.5           & 0.2         & 10-6         & 505            & 0.06        & 56.37    \\
NNCLR  & Cifar-100 & 100    & d=2048     & 0.5           & 0.3         & 10-6         & 518            & 0.074       & 55.02    \\
NNCLR  & Cifar-100 & 100    & d=2048     & 0.5           & 0.4         & 10-6         & 520            & 0.075       & 53.43    \\
NNCLR  & Cifar-100 & 100    & d=2048     & 0.5           & 0.15        & 10-7         & 492            & 0.048       & 56.19    \\
NNCLR  & Cifar-100 & 100    & d=2048     & 0.5           & 0.15        & 10-6         & 474            & 0.051       & 56.09    \\ \hline
\end{tabular}}
\caption{This table shows all the parameters, accuracies, rank scores, and mutual information values for the random parameter experiments shown in the main paper.}
\label{tab:random_list}
\end{table}
\clearpage
\section{Appendix Theoretical Details}
\label{sec:appendix_theory}

\subsection{High Level Intuition}
\label{sec:intuition}

 Higher dimensionality in $R$ is desirable because it counters the dimensional collapse effect discussed in \cite{jing2021understanding} and encourages a more diverse feature space. Lower $I(R;Z)$ is also desirable because it implies that the projection head is effective in removing uninformative features from the representation space. However, we prove through information theoretic bounds that increasing the dimensionality of $R$ causes a corresponding increase in $I(R;Z)$ thus necessitating a balance between the two for an ideal representation space.
This balancing act is illustrated in Figure~\ref{fig:intro} where an image is passed through an encoder $e(\cdot)$ to produce a representation space $R$ with 6 associated features. 3 features are target-relevant and 3 are uninformative. The feature space is associated with an eigenvalue distribution that indicates how relevant each feature is to the geometry of the representation space. Ideally, the eigenvalue distribution should capture just the target-relevant features; however, a higher dimensional space also captures uninformative features as shown in part a). To counter this, the projector should act as an information bottleneck \cite{tishby2000information} during training that projects the features into a lower dimensional space where only the target features are relevant. In part a), the distribution of eigenvalues remains the same after projection so the projection head does not remove spurious features from $R$ which corresponds to a high $I(R;Z)$. Part b) represents an ideal case where $R$ has sufficiently high dimensionality to capture mostly informative features while sufficiently low $I(R;Z)$ such that the projector guides the optimization towards target-relevant features. 

\begin{figure}[h!]
    \centering
    \includegraphics[width=\linewidth]{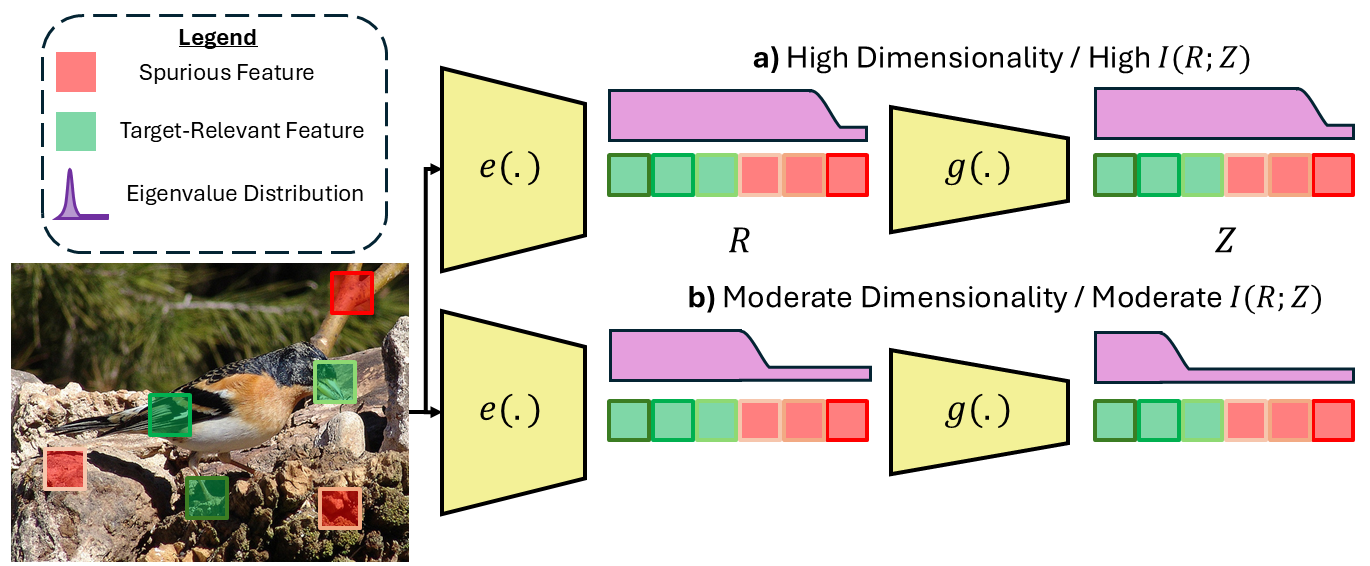}
    \caption{Assume there is an image with 3 task relevant features and 3 spurious features. The image is associated with a representation space $R$, projection space $Z$, and corresponding eigenvalue distributions for both. a) This is an example of $R$ and $Z$ with high dimensionality and high $I(R;Z)$. b) This is an example of $R$ and $Z$ that has moderate dimensionality and moderate $I(R;Z)$.}
    \label{fig:intuition}
\end{figure}

\subsection{Gaussian Mutual Info Derivation Details}
\label{sec:gaussian_derivation}

We follow the assumptions of Section \ref{sec:theory}. The following closed form equations are needed for this analysis:
\begin{align*}
I(R;Z) &= \frac{1}{2}(ln(|\Sigma_R|) +ln(|\Sigma_Z|) - ln(|\Sigma|)) \\
H(R) &= \frac{m}{2}ln(2\pi) + \frac{1}{2}ln(|\Sigma_R|) + \frac{m}{2} \\
ln(|\Sigma|) &= ln(|\Sigma_Z||\Sigma_R -\Sigma_{RZ}\Sigma_{Z}^{-1}\Sigma_{ZR}|) \\ 
ln(|\Sigma|) &= ln(|\Sigma_R||\Sigma_Z -\Sigma_{ZR}\Sigma_{R}^{-1}\Sigma_{RZ}|) \\
\end{align*}

Note that the $ln(|\Sigma|)$ derivation originated from Shur's complement formula. 

In this setting, $I(R;Z)$ can be rewritten as:

\begin{align*}
I(R;Z) &= \frac{1}{2}(ln(|\Sigma_R|) +ln(|\Sigma_Z|) - ln(|\Sigma_R||\Sigma_Z -\Sigma_{ZR}\Sigma_{R}^{-1}\Sigma_{RZ}|)\\
I(R;Z) &= \frac{1}{2}(ln(|\Sigma_R|) +ln(|\Sigma_Z|) - ln(|\Sigma_Z||\Sigma_R -\Sigma_{RZ}\Sigma_{Z}^{-1}\Sigma_{ZR}|)\\
\end{align*}

Using the law of logarithms, we can simplify this equation into:

\begin{align*}
I(R;Z) &= \frac{1}{2}(ln(|\Sigma_Z|) - ln(|\Sigma_Z -\Sigma_{ZR}\Sigma_{R}^{-1}\Sigma_{RZ}|)\\
I(R;Z) &= \frac{1}{2}(ln(|\Sigma_R|) - ln(|\Sigma_R -\Sigma_{RZ}\Sigma_{Z}^{-1}\Sigma_{ZR}|)\\
\end{align*}

This results in the form described in the main paper as: 
\begin{align*}
I(R;Z) = \frac{1}{2}(ln(|\Sigma_Z|) - ln(|Var(Z|R)|)) = \frac{1}{2}(ln(|\Sigma_R|) - ln(|Var(R|Z)|)) \\
\end{align*}

\begin{figure}[h!]
    \centering
    \includegraphics[width=\linewidth]{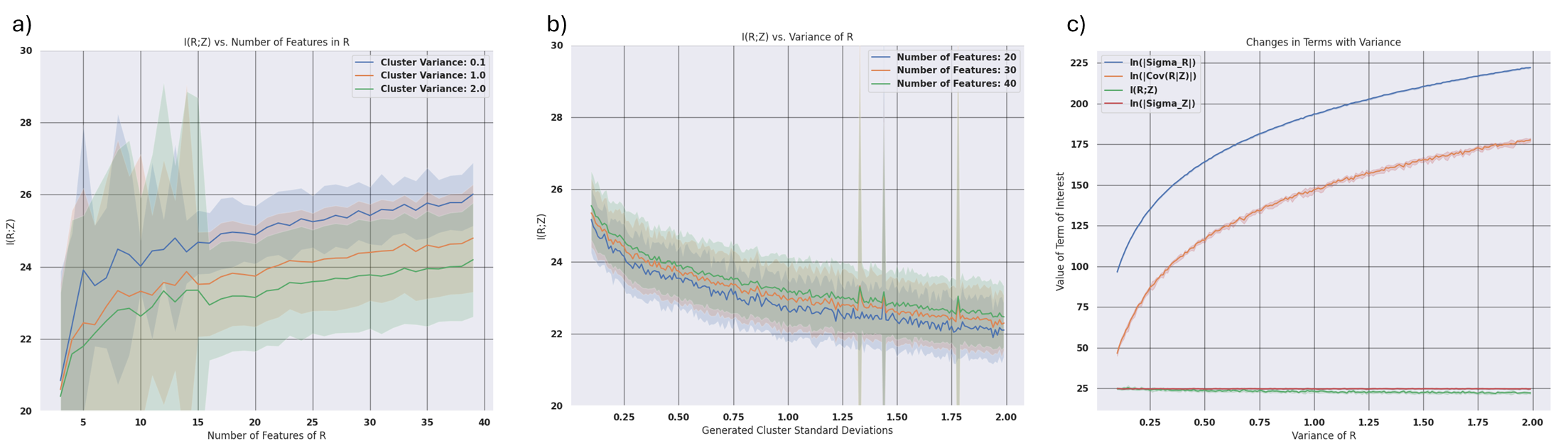}
    \caption{In a) and b), we show the $I(R;Z)$ curves as the number of features is varied while the cluster variance is kept constant. In c), we decompose each of the individual terms that make up $I(R;Z)$ and plot how their values vary during this same simulation .}
    \label{fig:terms}
\end{figure}

We further analyze the specific terms that make up this equation in Figure \ref{fig:terms}. In parts a) and b), the $I(R;Z)$ curves from the main paper are repeated. In part c), each of the terms that make up $I(R;Z)$ are analyzed the number of features are fixed and the sample variance is increased. $ln(|\Sigma_R|)$ and $ln(|Var(R|Z)|)$ increase as the variance increases. However, $ln(|Var(R|Z)|)$ increases at a comparatively faster rate. This happens because $ln(|\Sigma_Z|)$ does not change in value. The end result is a reduction in mutual information  which shows that $Z$ is not able to preserve the variance in $R$ under the conditions of its projection.

\subsection{Gaussian Simulation Details}
\label{sec:gaussian_sim}

In Figure \ref{fig:simulation}, we perform a detailed simulation on data generated from a Gaussian distribution. To generate this data, the make blobs dataset from the sklearn library\cite{pedregosa2011scikit} is used. This library generates Gaussian isotropic clusters that are intended for clustering problems. However, for our purposes it acts as a reliable generator of Gaussian distributed data. The cluster labels of this dataset are not used in any capacity for our experiments to conform to the SSL setting. This dataset has the following parameters:

\begin{enumerate}
    \item n\_samples: We set this to 1000 for all experiments.
    \item n\_features: We set this based on the features required for the simulation of interest. 
    \item centers: This is set to 5 for all experiments. This describes the number of clusters generated.
    \item cluster\_std: This is the parameter we vary to control the variance of the generated data.
    \item random\_state: This can set the initial random seed for the generation. We do not set this parameter. This is an intentional design choice so that we generate a slightly different version of the dataset on every iteration of the simulation. We take the average and standard deviation of 100 simulations for every set of parameters that we use in our experiments.
\end{enumerate}

\subsection{Neural Network Simulation}
\label{sec:network_sim}

In the main paper, PCA is used as a general projection between $R$ and $Z$ for the purposes of modeling the interaction between a space and its projection without having to deal with the nuances of training neural networks. However, the projector can also be replaced with a neural network and trained with either the NCE (SimCLR) or VICReg loss. We show that even in this setting the same general trends hold.

For this experiment, synthetic gaussian data is generated in the manner described in Section \ref{sec:gaussian_sim}. For training, a small MLP is used composed of 5 layers and 20 hidden units per layer, followed by a small projector with 2 layers and 5 hidden units per layer to output a dimension of size 5. The generated data has 25 features and a cluster standard deviation of .01. It is trained for 1000 epochs with the NCE (SimCLR) or VICReg loss. In this setting, augmentations are generated by adding randomly distributed Gaussian noise with a standard deviation of 0.5 to the generated data. During training, $I(R;Z)$ is measured for every epoch where $R$ is the original generated data and $Z$ is the output of the neural network. This value is computed using the closed form $I(R;Z)$ for gaussian distributed data. The Adam optimizer is used for these experiments with a learning rate of .0001 and a $\beta$ of 0.9 to 0.999.

\begin{figure}[h!]
    \centering
    \includegraphics[width=\linewidth]{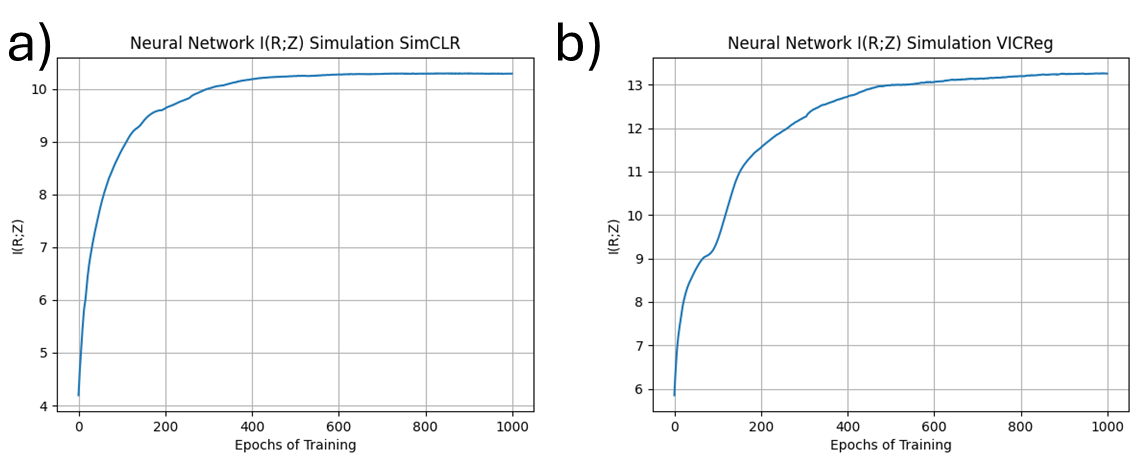}
    \caption{We show the $I(R;Z)$ curves across epochs of training for a gaussian dataset trained on a) SimCLR and b) VICReg.}
    \label{fig:network_sim}
\end{figure}

Figure~\ref{fig:network_sim} shows that the neural network simulation of our data exhibits the same trends as the PCA experiments found in the main paper. At the start of training, $I(R;Z)$ increases and gradually plateaus by the end of training. Additionally, the dimension contrastive strategy VICReg approaches a higher $I(R;Z)$ than that of the sample contrastive strategy SimCLR.

\subsection{Info Theoretic Bounds}
\label{sec:info_theory}

The upper bound on $I(R;Y)$ described originated from a derivation performed in \cite{ouyang2025projection}. The exact details of these bounds can be found in the original paper. Below we show a complete derivation of the bound described in equation~\ref{eq:info_gaussian}. The original bound is described as:

\begin{align*}
I(Y;R) \leq I(Y;Z) - I(R;Z) + H(R)
\end{align*}

The approximation $I(Y;Z) = G$ is used in the main paper which results in the following bound: 

\begin{align*}
I(Y;R) \leq G - I(R;Z) + H(R)
\end{align*}

We substitute in the equation $\frac{1}{2}(ln(|\Sigma_Z|) - ln(|Var(Z|R)|))$ for $I(R;Z)$ and $H(R) = \frac{m}{2}ln(2\pi) + \frac{1}{2}ln(|\Sigma_R|) + \frac{m}{2}$. This results in the bound:

\begin{align*}
I(Y;R) \leq G - \frac{1}{2}(ln(|\Sigma_Z|) - ln(|Var(Z|R)|)) + (\frac{m}{2}ln(2\pi)+ \frac{1}{2}ln(|\Sigma_R|) + \frac{m}{2})
\end{align*}

A simplification of terms results in the bound shown in the main paper as:

\begin{align*}
I(Y;R) \leq G + \underbrace{\frac{1}{2}(ln(|\Sigma_R|)-ln(|\Sigma_Z|))}_{K(Both)} + \underbrace{\frac{1}{2}ln(|Var(Z|R)|)}_{V(I(R;Z))} + \underbrace{\frac{m}{2}(ln(2\pi)+1)}_{D(H(R))}
\end{align*}

Note that for discrete random variables, Shannon's entropy is strictly non-negative. However, for continuous variables (such as multivariate Gaussians), the differential entropy can potentially be negative if the variance of the distribution is small. This could potentially lead to a mathematical contradiction where $I(Y;R) < I(Y;Z)$. However, as discussed in Section~\ref{sec:related}, SSL algorithms are optimized to increase the entropy of the representation space. In the case of Gaussian distributions, this corresponds to increasing the variance of the corresponding random variable. For this reason, we assume that the variance at the end of the training is sufficiently large enough to avoid this contradiction. We also note that our analysis results in a dimension term $m$. We interpret this term to reflect the number of non-redundant dimensions as reflected by metrics such as effective rank. Therefore, this term can increase or change throughout training as the dimensionality or variance of the space projects into new dimensions.

The SSL training dynamics discussed in Section~\ref{sec:dynamics} leads to the emergence of a dynamical system where increasing one of these terms can potentially limit the growth of the others. For example, if $H(R)$ increases via feature decorrelation, then $D$ will increase due to a greater number of features $m$. This, in turn, causes feature decorrelation in $Z$, as discussed in our previous dynamics, which causes $I(R;Z)$ to increase and $V$ to decrease which limits the upper bound in equation~\ref{eq:info_gaussian}.   
Additionally, $K$ is limited in this setting due to both of its terms increasing together. In the setting where $H(R)$ increases due to sample uniformity, $D$ is fixed in the number of dimensions which acts as a bound on how large $H(R)$ can grow. In contrast, $K$ and $V$ increase due to an increase in the variance of $R$ without the projection $Z$ having a corresponding increase in variance which lowers $I(R;Z)$. 
\clearpage
\section{Appendix Analytical Details}
\label{sec:appendix_analysis}

\subsection{Limitations and Broader Impact}
\label{sec:limit}
Our work focuses on finding an optimal point between $H(R)$ and $I(R;Z)$ and analyzing the training dynamics that influence both terms. However, the notion of an ``ideal`` optimal point is difficult to prove with respect to a given data setting. Ideally, there should be a derived bound or ratio between $H(R)$ and $I(R;Z)$ that we can claim with high probability to correspond to an ``ideal`` optimal relationship between the two terms. Despite this limitation, the broader impact of this work is that it provides a general framework to develop SSL algorithms across diverse fields such as medicine \cite{kokilepersaud2023clinically}, seismology \cite{kokilepersaud2022volumetric}, and autonomous driving \cite{kokilepersaud2023exploiting}. Therefore, it provides an avenue for potential growth of machine learning solutions in a wide variety of fields. We are unaware of any negative societal impacts directly caused by our work.

\subsection{VICReg vs. SimCLR Comparison}
\label{sec:vicreg_simclr}
\begin{figure}[h!]
    \centering
    \includegraphics[width=\linewidth]{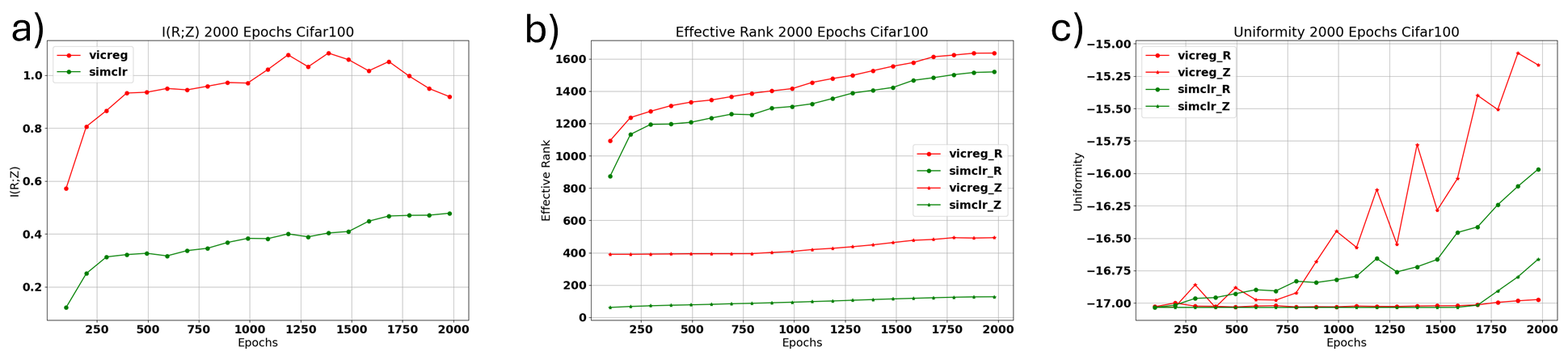}
    \caption{We train models with two different SSL methods for 2000 epochs and then analyze changes in a) $I(R;Z)$, b) effective rank between $R$ and $Z$, and c) uniformity between $R$ and $Z$.}
    \label{fig:2000_simulation_comparison}
\end{figure}

The \texttt{AdaDim} methodology is based on the premise that VICReg better promotes higher $H(R)$ and the NCE loss used in SimCLR promotes lower $I(R;Z).$
These same dynamics are observed in a real SSL setting in Figure~\ref{fig:2000_simulation_comparison} where a ResNet-50 model is trained for 2000 epochs on Cifar-100 \cite{krizhevsky2009learning} using the VICReg and SimCLR SSL methods. In part a), both methods have an increase in $I(R;Z)$, but it occurs at a slower rate for SimCLR. In part b), the overall dimensionality of the dataset increases across all training epochs for $R$, but begins to plateau at the end of training, corresponding to the end of the feature decorrelation stage. $Z$ exhibits this same behavior, but plateaus much more noticeably throughout training which contributes to the plateauing effect of $I(R;Z)$. For both $R$ and $Z$, the overall dimensionality is lower for SimCLR than for VICReg. In part c), $R$ and $Z$ have a similar uniformity for both methods at the start of training, but significantly diverge from each other by the end of training.

\begin{figure}[h!]
    \centering
    \includegraphics[width=\linewidth]{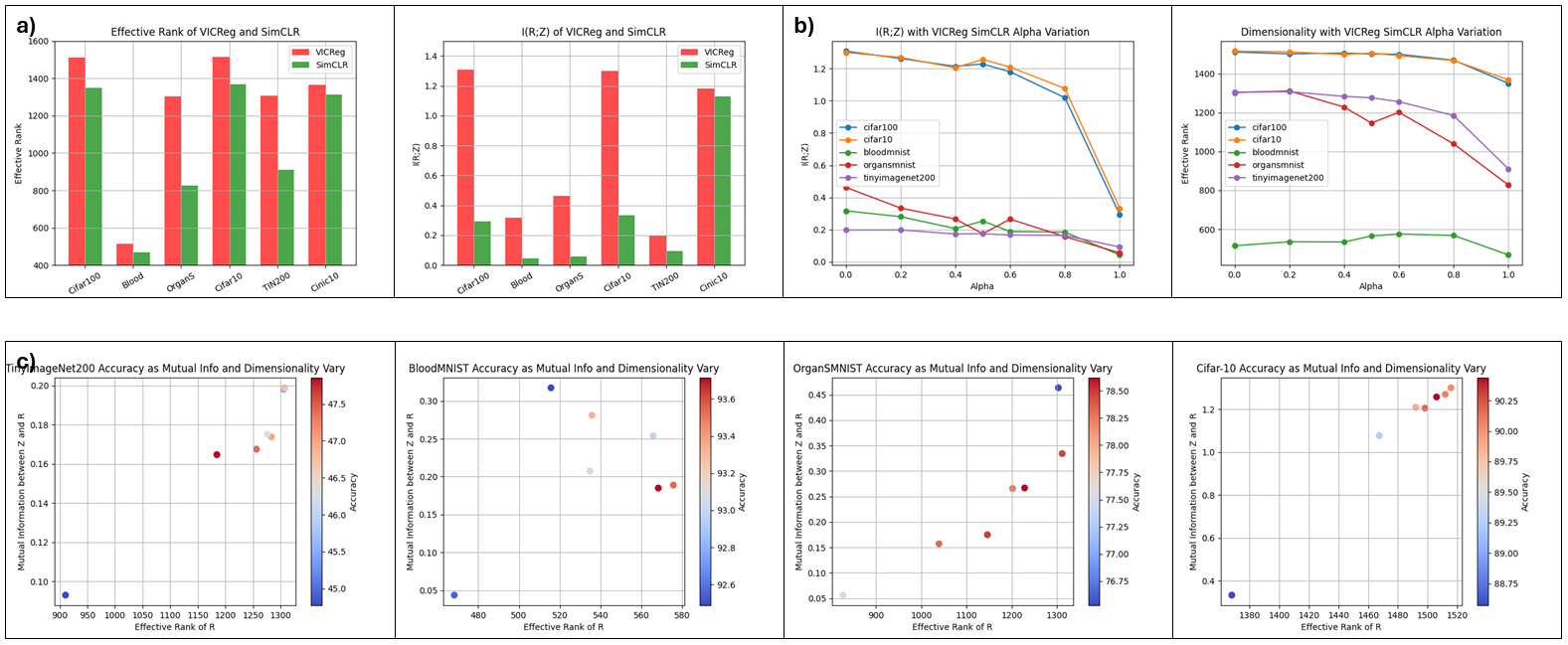}
    \caption{a) We compare the effective rank and $I(R;Z)$ between the representation of the test set for different datasets trained  on VICReg and SimCLR. b) We show how the effective rank and $I(R;Z)$ vary after the introduction of the $\alpha$ parameter for each dataset. c) We show how performance varies as a function of $I(R;Z)$ and $H(R)$ for a variety of datasets.}
    \label{fig:dim_mutual_info_datasets}
\end{figure}

These observed trends for SimCLR and VICReg hold for a wide variety of datasets in parts a) and b) of Figure \ref{fig:dim_mutual_info_datasets}. In part a), at the end of training for 6 different datasets, the $H(R)$ and $I(R;Z)$ values of VICReg are higher than those of SimCLR. In part b), these trends are analyzed over the course of manually setting the $\alpha$ parameter over the course of training from 0 to 1 in increments of 0.2. It is observed that as the optimization changes from VICReg ($\alpha = 0$) to SimCLR ($\alpha = 1$), the $I(R;Z)$ and $H(R)$ values monotonically decrease. 

\subsection{Generation of $H(R)$ vs $I(R;Z)$ Plots}
\label{sec:trade-off-gen}

In part c) of Figure \ref{fig:dim_mutual_info_datasets}, empirical results demonstrate the existence of an optimal point between high $H(R)$ and low $I(R;Z)$. The plots are generated by training each respective dataset with manually chosen $\alpha$ parameters within the \texttt{AdaDim} framework. In this case, manual means that the adaptive $\alpha$ computation is removed and a specific value of $\alpha$ from 0 to 1 is kept constant during training with the $\gamma$ parameter set to 0. These $\alpha$ values are $\alpha = [0,0.2,0.4,0.5,0.6,0.8,1.0]$. The logic behind this setup is to provide a controlled system where we can gradually transition between a feature decorrelation based loss and a  sample uniformity based loss.  The plot in part a) of Figure \ref{fig:intro} was generated in a similar manner. However, in this case, the $\alpha$ increments are reduced to 0.05 In this way, we can more easily observe the emergence of the $H(R)$ and $I(R;Z)$ balance regions without needing to train as many models as is done in Figure~\ref{fig:random}.

More plots are provided for more datasets in part c) of Figure \ref{fig:dim_mutual_info_datasets}. Note that potentially more models need to be generated in order for the balance trend to be more salient for specific datasets. However, for many of these datasets, such as TinyImageNet, OrganSMNIST, and Cifar-10, the trend is clear that performance depends on a balance between $H(R)$ and $I(R;Z)$. 

We also have experiments in which the $H(R)$ and $I(R;Z)$ plots are generated by completely random parameters. This is shown in Figure~\ref{fig:random}. In these cases, the trend of a balancing point is not as smooth, but it is clear that there exists a cluster of points around specific $H(R)$ and $I(R;Z)$ values that perform better than those points outside the discussed region. The exact parameters used to generate these graphs are provided in Table~\ref{tab:random_list}. 

The same list of results is used to compute the correlation experiments in Table~\ref{tab:corr}. In this table, we compute the results for a specified number of models and their corresponding results. We then compute the pearson correlation coefficient between the list of results for each model and the metric of interest that includes $H(R)$, $I(R;Z)$, or the ratio between them. In this case, the ratio is computed as simply $\frac{I(R;Z)}{H(R)}$. In the table, we report the magnitude of the correlation coefficient as we are only interested in the existence of a linear relationship between both terms, not whether it is positively or negatively correlated. In general, for all metric parameters, it is dataset dependent as to whether the correlation is positive or negative. This agrees with the analysis of our paper since each dataset has its own dimensionality characteristics and may favor a specific relationship between $H(R)$ and $I(R;Z)$.

\begin{table}[h!]

\begin{tabular}{@{}ccccccc@{}}
\toprule
\multicolumn{7}{c}{Magnitude of Correlation with Performance Across Trained Models}              \\ \midrule
Method & Dataset         & Epochs & \# of Models & H(R)         & I(R;Z)        & Ratio         \\ \midrule
SimCLR & Cifar100        & 100    & 15           & .082          & .323          & \textbf{.462} \\
VICReg & Cifar100        & 100    & 15           & .013          & \textbf{.772} & .751          \\
NNCLR & Cifar100        & 100    & 15           & .206          & .229 & \textbf{.337}        \\
All-ResNet18 & Cifar100        & 1000    & 11           & .029          & .372 & \textbf{.375}        \\
SimCLR & Cifar100        & 400    & 10           & .557          & .543          & \textbf{.625} \\
VICReg & Cifar100        & 400    & 10           & .351          & .875          & \textbf{.894} \\
SimCLR & TinyImageNet200 & 400    & 10           & \textbf{.534}          & .507          & .521 \\
SimCLR & Cinic-10        & 400    & 10           & .029          & .323          & \textbf{.421} \\
SimCLR & Cifar-10        & 400    & 10           & \textbf{.873} & .841          & .833          \\
SimCLR & OrganSMNIST     & 400    & 10           & .0024         & .435          & \textbf{.442} \\ \bottomrule
\end{tabular}
\caption{This table shows the pearson correlation coefficient between the performance of a set of SSL models trained with different hyperparameters on a specific dataset and the effective rank ($H(R)$), $I(R;Z)$, and the ratio between them. }%
  \label{tab:corr}
\end{table}

\subsection{Discussion of Relationship with Literature}

One surprising observation of this paper is that it opposes a variety of recent works \cite{agrawal2022alpha,thilak2023lidar,garrido2022duality}. In these papers, the authors argue that some measure of dimensionality can be used as an unsupervised surrogate of representational quality. In other words, higher dimensionality should correspond to a better performing model on potentially any downstream task. However, our work suggests that both $H(R)$ and $I(R;Z)$ should be considered for an unsupervised assessment of model quality. However, our result is not surprising when we consider how these works justify their conclusions. For example, \cite{garrido2023rankme} based their rank estimates off of pre-trained ImageNet models which may not reflect the dynamics of training an SSL method on any given dataset. \cite{thilak2023lidar} showed a wide range of coefficient correlation values (0.2 - 0.8) between different dimensionality based metrics and performance values derived from various sources. This suggests that in some settings dimensionality is a good surrogate for performance while in others $I(R;Z)$ needs to be considered. This corresponds to the dynamics discussed in this paper, where the best performing model is often not the one with the highest dimensionality. It is the one that reaches a suitable intermediate point between both $H(R)$ and $I(R;Z).$ Our work suggests that future unsupervised estimators of representational quality should have some mechanism to detect this balance between the two terms of interest.

We also note that \cite{garrido2022duality} discussed conditions under which both sample and dimension contrastive approaches are equivalent from an optimization perspective. However, the authors of this work also acknowledge that neither approach can be used interchangeably. More recent work \cite{shwartz2023information} identified differences in the entropy characteristics of both approaches and suggested that it may be possible to devise methods that are able to take advantage of the characteristics of both. Therefore, our work can also be understood from the perspective of identifying the underlying dynamics where both approaches contribute meaningfully to SSL optimization.

\subsection{Manual $\alpha$ Usage}
\label{sec:alpha_analysis}
\begin{table}[h!]
\centering
\scalebox{.8}{
\begin{tabular}{ccccccccc}
\hline
       &       & \multicolumn{7}{c}{Dataset}                                                                                  \\ \cline{3-9} 
Method & Alpha & Cifar100       & Cifar10        & TinyImageNet200 & Cinic10        & Blood & OrganS         & iNat21         \\ \hline
SimCLR & N/A   & 64.00          & 88.59          & 44.78           & 78.54          & 92.54 & 77.67          & 23.96          \\
VICReg & N/A   & 64.70          & 90.02          & 45.54           & 78,25          & 92.48 & 76.50          & 24.24          \\ \hline
AdaDim & 0.2   & 65.18          & 90.07          & 46.75           & 78.27          & 93.36 & 78.41          & -              \\
AdaDim & 0.4   & 66.15          & 90.18          & 47.00           & 78.57          & 93.04 & 78.46          & -              \\
AdaDim & 0.5   & 66.53 & 90.43          & 46.26           & 79.35          & 92.98 & 78.50          & 24.56 \\
AdaDim & 0.6   & 66.11          & 89.87          & \textbf{48.06}  & \textbf{79.58} & 93.56 & 78.23          & -              \\
AdaDim & 0.8   & 66.32          & 89.25          & 47.83           & 78.54          & \textbf{93.71} & 78.26          & -              \\ \hline
AdaDim & Ada   & \textbf{66.90}          & \textbf{90.72} & 47.81           & 79.53          & 92.86 & \textbf{78.55} & \textbf{24.81}              \\ \hline
\end{tabular}}
\caption{This shows the performance of AdaDim under different $\alpha$ parameters on several different datasets. $\gamma$ is set to 0 for this study.}
\label{tab:alpha}
\end{table}
In Table \ref{tab:alpha}, an ablation study is performed where the $\alpha$ parameter is varied and the $\gamma$  parameter is set to 0. In this table, we use the symmetric augmentation scheme of \cite{da2022solo} and pre-train each method for 400 epochs on a ResNet-50 model. This specific training setting is used as it allows us to better isolate changes in performance due to the specific optimization objectives and not due to other considerations such as the applied augmentation. We find that our adaptive methodology without $\gamma$ either outperforms or is consistent with the best result that we obtain from manually choosing a hyperparameter for $\alpha$. This highlights the importance of adaptively shifting between losses over the course of training to match the dynamics of SSL training.

\begin{figure}[h!]
    \centering
    \includegraphics[width=\linewidth]{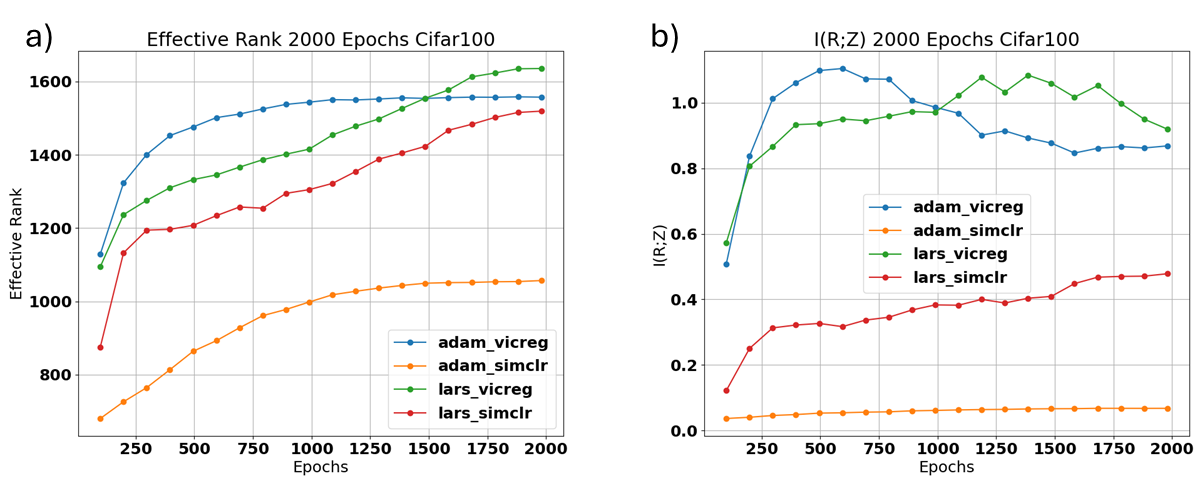}
    \caption{We show how a) $H(R)$ and b) $I(R;Z)$ varies for both SimCLR and VICReg under different optimization settings.}
    \label{fig:optimization_procedure}
\end{figure}

\subsection{Variation in Optimization Procedure}
\label{sec:optimizer}

In Figure \ref{fig:optimization_procedure}, we vary the optimization setting for both SimCLR and VICReg. It is observed that the effective rank and $I(R;Z)$ curves have similar trends for both the adam and lars optimizers. However, the difference is that for the adam optimizer, the effective rank has a more pronounced upper limit on the values it can reach. Additionally, for $I(R;Z)$, the adam trained optimizer begins to decrease or plateau quicker. This result highlights that the trends of this paper are general, but its exact manifestation across training will vary based on the setup of the experiment.

\subsection{$\gamma$ Analysis}

\begin{figure}[h!]
    \centering
    \includegraphics[scale=0.4]{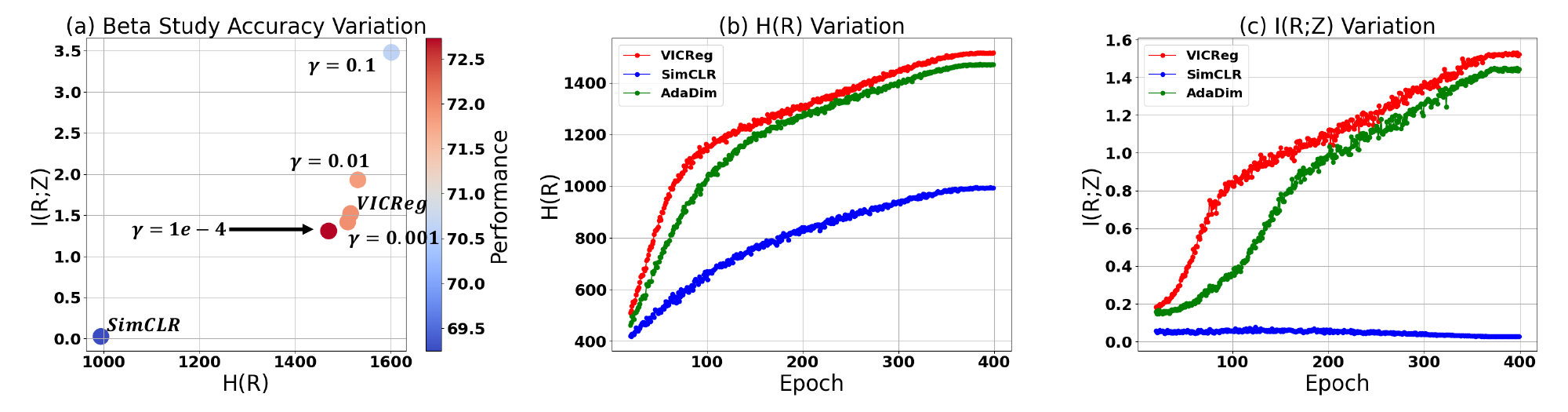}
    \caption{a) This figure shows how performance, $H(R)$, and $I(R;Z)$ change as $\gamma$ is varied. b) This figure shows how the effective rank of AdaDim varies compared to baseline methods. c) This figure shows how the $I(R;Z)$ for AdaDim varies compared to baseline methods.}
    \label{fig:beta_analyze}
\end{figure}

In part a) of Figure \ref{fig:beta_analyze}, we perform an analysis of varying $\gamma$. We find that increasing $\gamma$ causes an increase in both $I(R;Z)$ and $H(R)$ while decreasing $\gamma$ causes the opposite effect. It is interesting to note that the same trend of an ideal balance emerges through the choice of the $\gamma$ parameter. In this case, performance is maximized when choosing $\gamma$ that results in more intermediate values between both $H(R)$ and $I(R;Z)$. This illustrates the potential importance of tuning $\gamma$ for specific data settings as each dataset will have a specific balance between $H(R)$ and $I(R;Z)$ that is optimal.

 In parts b) and c), we compare the growth of  $H(R)$ and $I(R;Z)$ over the course of training a ResNet-50 model for 400 epochs on Cifar-100. We find that the adaptive regularization technique of our method leads to \texttt{AdaDim} reaching a trajectory that attempts to balance between both terms when compared with the more biased trajectories of SimCLR and VICReg. 

We also note that the trajectory of AdaDim follows a similar trajectory to that of VICReg. In the design of our method, VICReg is used much more heavily at the start of training which is when much of the structure of the representation space is learnt. However, the gradual introduction of the NCE loss of SimCLR is an important aspect gradually introduce sample uniformity at the end of training. This importance is reflected in the significant performance improvements throughout the paper even if its introduction only lowers the corresponding H(R) and I(R;Z) values slightly.

\subsection{Vision Transformer Experiments}

\begin{table}[]
\centering
\begin{tabular}{ccc}
\hline
\multicolumn{3}{c}{ViT Cifar-100 Comparison} \\ \hline
Method      & Swin-Small \cite{liu2021swin}     & ViT-Large \cite{dosovitskiy2020image}     \\ \hline
SimCLR      & 53.91          & 63.92         \\
VICReg      & 56.83          & 63.91         \\
AdaDim      & \textbf{58.07}          & \textbf{66.35}         \\ \hline
\end{tabular}
\caption{This table shows the the performance of AdaDim under a vision transformer setting \cite{liu2021swin,dosovitskiy2020image}.}
\label{tab:vit_compare}
\end{table}

In Table~\ref{tab:vit_compare}, we provide comparisons of AdaDim when trained with a vision transformer backbone. Specifically, we perform a comprehensive study of a smaller scale transfomer \cite{liu2021swin} and a larger scale ViT model \cite{dosovitskiy2020image}. We train these networks on Cifar-100 for 400 epochs, with an AdamW optimizer, batch size of 256, learning rate of 2e-4, weight decay of 0.05, and betas of 0.9 to 0.95. For these experiments, we use the asymmetric augmentation setting. We find that AdaDim out performs standard VICReg and SimCLR baselines for both models. The reason for this improvement may be due to specific properties regarding transformers. Specifically, transformers benefit from longer training times compared to traditional CNNs. When training with more epochs, there is a greater ability for the model to go through the training dynamics dynamics discussed in this paper. It is possible that AdaDim is advantageous in this setting due to providing explicit regularization based on these emergent dynamics. 

\subsection{Comparison Against Alternate Training Strategies}
\label{sec:alternate_strategies}
\begin{table}[]
\scalebox{0.75}{
\begin{tabular}{@{}ccccccccc@{}}
\toprule
\multicolumn{9}{c}{AdaDim Standardized Hyperparameter Ablation Study}                                                                                                                 \\ \midrule
Method                          & $\alpha$ Type  & Cifar100        & Cifar10 & TinyImageNet200 & Cinic10        & Blood          & OrganS         & iNat21         \\ \midrule
SimCLR \cite{chen2020simple}                         & N/A                         & 64.00           & 88.59   & 44.78           & 78.54          & 92.54          & 77.67          & 23.96          \\
VICReg  \cite{bardes2021vicreg}                        & N/A                         & 64.70           & 90.02   & 45.54           & 78.25          & 92.48          & 76.50          & 24.24          \\
SimCLR + $\lambda$ \cite{ouyang2025projection} & N/A                         & 64.37           & 88.00   & 45.54           & 76.96          & 92.86          & 77.98          & 23.51          \\
VICReg + $\lambda$ \cite{ouyang2025projection} & N/A                         & 64.54           & 89.77   & 45.83           & 78.47          & 92.43          & 77.16          & 24.01          \\ \midrule
SimCLR +VICReg                  & $\alpha$ = 0.5 & 66.53           & 90.43   & 46.26           & 79.35          & 92.86          & 78.50          & 24..56         \\
SimCLR +VICReg                  & cosine                      & 65.78           & 88.85   & 45.45           & 78.87          & 92.57          & 78.46          &        -        \\
SimCLR +VICReg                  & linear                      & 66.99           & 89.53   & 45.94           & 78.60          & 92.07          & 78.61          &         -       \\ \midrule
AdaDim ($\alpha =$ Ada)                        & Ours                        & 66.90  & 90.72   & 47.81          & 78.55          & 93.10    & 78.55          & -         \\ 

AdaDim ($\alpha =$ Ada, $\beta=$ Ada)                        & Ours                        & \textbf{67.15}  & \textbf{90.81}   & \textbf{48.24}           & \textbf{79.53}          & \textbf{93.24}          & \textbf{79.19}          & \textbf{24.81}          \\\bottomrule
\end{tabular}}{%
  \caption{This table shows an ablation study of performance across a variety of datasets when varying aspects of the construction of the AdaDim loss.}
  \label{tab:sota_ablation}
}
\end{table}
In Table~\ref{tab:parameter_var}, we provide an ablation study of the impact of the different components of our method and hypothetical ways our method could be applied. This includes strategies such as a fixed $\beta$ parameter during training and a heuristic scheduler of the $\alpha$ parameter. Possible schedules include cosine or linear scaling from 0 to 1. To further validate the importance of adaptive scaling of both the $\alpha$ and $\beta$ terms, we provide a comprehensive study of these terms under a fixed hyperparameter setting in Table~\ref{tab:sota_ablation}. The experiments in this table are setup such that each method has the exact same hyperparameters to ensure that any deviations in performance are caused solely by the optimization objective. This includes 400 epochs of training with a ResNet-50 model, the symmetric augmentation scheme of \cite{da2022solo}, a LARS optimizer, learning rate of 0.4, and a batch size of 256 for all experiments. Again, we observe that our method that makes use of a dimensionality adaptation technique out performs heuristic schedules as well as methods based on fixed regularization of $I(R;Z)$ \cite{ouyang2025projection} across a wide variety of data settings.

\begin{table}[]
\begin{tabular}{ccc}
\hline
\multicolumn{3}{c}{AdaDim Batch Size Comparison Experiments}  \\ \hline
Batch Size & ImageNet-100 + LARS & Cifar-100 + SGD + Momentum \\ \hline
128        & 79.30               & -                          \\
256        & 79.64               & 74.31                      \\
512        & 80.06               & 71.67                      \\
1024       & -                   & 71.02                      \\ \hline
\end{tabular}
\caption{In this table, we compare the performance as batch sized is varied for two different experimental settings. Both methods are trained with a ResNet-18 encoder. ImageNet-100 experiments are trained for 400 epochs while Cifar-100 experiments  are trained for 1000 epochs. For these experiments, the $\gamma$ parameter is kept fixed at 1e-4. ImageNet-100 uses the baseline AdaDim strategy, while Cifar-100 is trained with the AdaDim + momentum method.}
\label{tab:batch_size_ablation}
\end{table}

\subsection{Batch Size Discussion}

In the main paper, we conduct all our experiments with small batch sizes of 128 or 256 depending on the context of the experiment. In all cases, we achieve state of the art performance without necessitating large batch sizes as discussed in \cite{chen2020simple}. However, we also wanted to assess how our method would perform in large batch settings. We perform this analysis in Table~\ref{tab:batch_size_ablation} where we compare the performance of our method in two divergent settings. This includes the baseline hyperparameter setting with ImageNet-100 and the expanded hyperparameter setting with a momentum encoder trained on Cifar-100. In the case of ImageNet-100 with the LARS optimizer, we find that performance improves with larger batch sizes. However, with Cifar-100 and the SGD optimizer, performance degrades with the introduction of larger batch sizes. We believe the reaction to batch size is largely a matter of identifying optimal optimization settings for the large batch setting. For example, ImageNet-100 possibly improved due to the LARS optimizer being better suited to scaling optimization with respect to the large batch setting. However, for the SGD setting, further tuning of learning rates and parameters such as $\gamma$ may be needed for effective downstream performance on large batches. Additionally, the batch setting improvements may also be dataset dependent with ImageNet-100 benefiting much more from large batches than Cifar-100. This may be the case as most batch size results in SSL papers \cite{chen2020simple,chen2020improved}
are benchmarked on large scale datasets such as ImageNet, rather than Cifar-100. Regardless, we see the small batch size state of the art performance of our method as a major advantage compared to the resource intensive nature of other methods.

\subsection{Training Variation Analysis}

\begin{figure}[h!]
    \centering
    \includegraphics[width =\columnwidth]{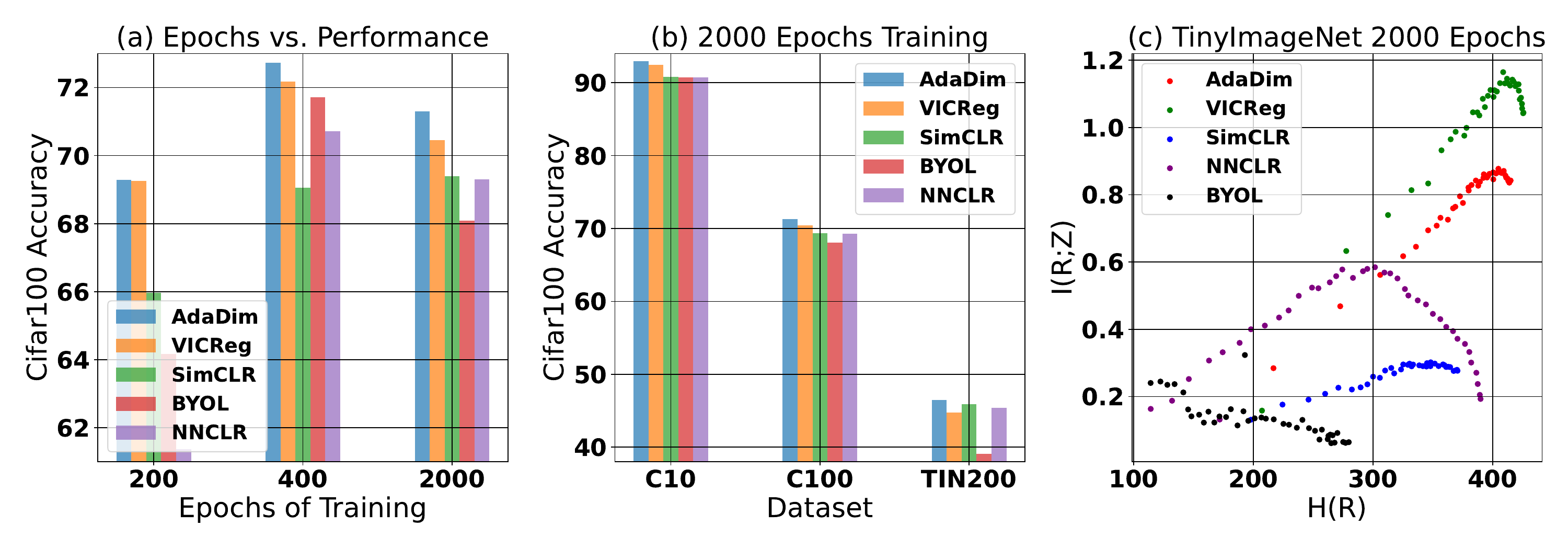}
    \caption{a) This figure demonstrates the performance of AdaDim under different amounts of training epochs on Cifar-100. b) This figure shows how AdaDim performs under a long training setting  for Cifar-10 (C10), Cifar-100 (C100), and TinyImageNet200 (TIN200) c) This shows the $H(R)$ and $I(R;Z)$ trajectory over 2000 epochs on 4 different algorithms.}
    \label{fig:training_epochs}
\end{figure}

We also analyze performance under different amounts of training epochs in part a) of Figure~\ref{fig:training_epochs} and on a long training regime across 3 datasets in part b). This indicates that AdaDim is adapts to SSL dynamics as training progresses. This is further exemplified by the  $H(R)$ and $I(R;Z)$ trajectory plot in part c) of Figure~\ref{fig:training_epochs} where AdaDim effectively balances between both terms to a greater degree than other methods.

We also note that VICReg saturates in performance much quicker than AdaDim and the rate of performance improvement across training increases much more substantially for AdaDim in Table \ref{tab:epochs}. We also note that from the perspective of matched compute budgets, AdaDim is still viable. In the case where we assume that VICReg is 20\% faster, this corresponds to AdaDim having a budget of 800 epochs while VICReg is able to utilize 1000 epochs. AdaDim at 800 epochs is still 2\% performance accuracy better than VICReg at the end of training for 1000 Epochs. This statement is also true at 500 epochs where AdaDim at 500 epochs has a better performance than VICReg at 1000 epochs of training. This behavior is mainly due to the emergence of dynamics during longer training settings that our AdaDim algorithm is able to adapt to. However, traditional SSL strategies like VICReg saturate much earlier as they lack this adaptive mechanism.

\begin{table}[h!]
\begin{tabular}{@{}ccc@{}}
\toprule
Epoch                     & VICReg                     & AdaDim                     \\ \midrule
\multicolumn{1}{|c|}{200} & \multicolumn{1}{c|}{66.25} & \multicolumn{1}{c|}{67.46} \\ \midrule
\multicolumn{1}{|c|}{300} & \multicolumn{1}{c|}{68.35} & \multicolumn{1}{c|}{69.73} \\ \midrule
\multicolumn{1}{|c|}{400} & \multicolumn{1}{c|}{69.70} & \multicolumn{1}{c|}{70.64} \\ \midrule
\multicolumn{1}{|c|}{500} & \multicolumn{1}{c|}{70.66} & \multicolumn{1}{c|}{71.97} \\ \midrule
\multicolumn{1}{|c|}{600} & \multicolumn{1}{c|}{71.28} & \multicolumn{1}{c|}{72.58} \\ \midrule
\multicolumn{1}{|c|}{700} & \multicolumn{1}{c|}{71.10} & \multicolumn{1}{c|}{73.43} \\ \midrule
\multicolumn{1}{|c|}{800} & \multicolumn{1}{c|}{71.36} & \multicolumn{1}{c|}{73.79} \\ \midrule
\multicolumn{1}{|c|}{900} & \multicolumn{1}{c|}{71.91} & \multicolumn{1}{c|}{73.83} \\ \midrule
1000                      & 71.70                      & \multicolumn{1}{l}{74.31}  \\ \bottomrule
\end{tabular}
\caption{In this table, we show the intermediate performance at different epochs for VICReg and AdaDim for 1000 epochs of training on ResNet-18 in the expanded hyperparameter settings.}
\label{tab:epochs}
\end{table}

\subsection{Standard Deviation Analysis}
\label{sec:stdev}

In general, our choice of only showing the single result in the tables of the main paper is due to the convention of previous SSL papers. However, the reason that this convention exists is because, in general, SSL strategies are benchmarked on well structured datasets and the process of pre-training across a large number of epochs on a large well made dataset followed by a robust fine-tuning step actually causes the variability between runs to be incredibly small. This trend is especially true when using the parameters of state of the art methods. To illustrate this,  we train 3 very different datasets: Cifar-100, TinyImageNet200, and BloodMNIST with 4 randomly generated seeds each using the same exact AdaDim+momentum training setup each time as discussed in the main paper. This is shown in Table~\ref{tab:stdev} We find that the variation between runs is less than .1\% accuracy for the well-curated Cifar-100 and TinyImageNet200 and less than .3\% for the more imbalanced and noisy medical blood dataset. This highlights the low variability between runs. 

\begin{table}[]
\begin{tabular}{@{}llll@{}}
\cmidrule(l){2-4}
\textbf{} & \textbf{Cifar100} & \textbf{TinyImageNet200} & \textbf{Bloodmnist} \\ \midrule
Average   & 75.4              & 54.725                   & 95.62               \\
STDEV     & 0.0424            & 0.063                    & 0.291       \\ \bottomrule
\end{tabular}
\caption{This table shows the performance variation between runs on 3 datasets across 4 random seeds.}
\label{tab:stdev}
\end{table}

\subsection{ImageNet Fine-Tuning Experiments}
\begin{table}[h!]
\centering
\scalebox{0.6}{
\begin{tabular}{cccccccccccc}
\cline{6-12}
\multicolumn{1}{l}{} & \multicolumn{1}{l}{} & \multicolumn{1}{l}{} & \multicolumn{1}{l}{} & \multicolumn{1}{l}{}   & \multicolumn{3}{c}{ImageNet Pre-Training}                                     & \multicolumn{4}{c}{ImageNet Fine-Tuning}                                            \\ \hline
Method               & M                    & P                    & Q                    & \multicolumn{1}{c|}{C} & 100 Epochs (R50) & 100 Epochs (ViT-S) & \multicolumn{1}{c|}{200 Epochs (R50)} & CUB200         & Food-101       & Stanford Dogs  & \multicolumn{1}{c|}{Oxford Pets} \\ \hline
Resa                 &    \checkmark                  &   \xmark                   &      \xmark                &      \checkmark                  & 70.80            & 67.98              & 71.59                                 & 54.98          & 73.21          & 69.71          & 86.10                            \\
AdaDim (ours)        &        \checkmark              &      \xmark                &     \xmark                 &    \xmark                    & \textbf{71.01}   & \textbf{68.31}     & \textbf{72.84}                        & \textbf{59.03} & \textbf{74.16} & \textbf{71.61} & \textbf{88.25}                   \\ \hline
\end{tabular}}
\caption{In this table, we present an ablation study of our method across different amounts of ImageNet pre-training on a ResNet-50 and ViT-Small backbone as well as in the context of fine-tuning to different datasets after pre-training on a ResNet-50 encoder for 100 epochs.}
\label{tab: imagenet_ablation}
\end{table}
%%%%%%%%%%%%%%%%%%%%%%%%%%%%%%%%%%%%%%%%%%%%%%%%%%%%%%%%%%%%

\newpage
\section*{NeurIPS Paper Checklist}

\begin{enumerate}

\item {\bf Claims}
    \item[] Question: Do the main claims made in the abstract and introduction accurately reflect the paper's contributions and scope?
    \item[] Answer: \answerYes{} % Replace by \answerYes{}, \answerNo{}, or \answerNA{}.
    \item[] Justification: We explicitly state our claims in bold in the introduction and then reinforce it throughout the paper.

\item {\bf Limitations}
    \item[] Question: Does the paper discuss the limitations of the work performed by the authors?
    \item[] Answer: \answerYes{} % Replace by \answerYes{}, \answerNo{}, or \answerNA{}.
    \item[] Justification: Limitations are discussed in Section~\ref{sec:limit}.

\item {\bf Theory assumptions and proofs}
    \item[] Question: For each theoretical result, does the paper provide the full set of assumptions and a complete (and correct) proof?
    \item[] Answer: \answerYes{} % Replace by \answerYes{}, \answerNo{}, or \answerNA{}.
    \item[] Justification: Theoretical proofs and assumptions are discussed in Section~\ref{sec:appendix_theory}.

    \item {\bf Experimental result reproducibility}
    \item[] Question: Does the paper fully disclose all the information needed to reproduce the main experimental results of the paper to the extent that it affects the main claims and/or conclusions of the paper (regardless of whether the code and data are provided or not)?
    \item[] Answer: \answerYes{} % Replace by \answerYes{}, \answerNo{}, or \answerNA{}.
    \item[] Justification: All experimental details are discussed in Section~\ref{sec:appendex_experiments}.

\item {\bf Open access to data and code}
    \item[] Question: Does the paper provide open access to the data and code, with sufficient instructions to faithfully reproduce the main experimental results, as described in supplemental material?
    \item[] Answer: \answerYes{} % Replace by \answerYes{}, \answerNo{}, or \answerNA{}.
    \item[] Justification: All experiments are performed on publicly available datasets and on open source repositories discussed in Section~\ref{sec:appendex_experiments}. While we do not release our codebase until after acceptance, using the provided repositories with the details and parameters in the supplementary will make our results reproducible.

\item {\bf Experimental setting/details}
    \item[] Question: Does the paper specify all the training and test details (e.g., data splits, hyperparameters, how they were chosen, type of optimizer) necessary to understand the results?
    \item[] Answer: \answerYes{} % Replace by \answerYes{}, \answerNo{}, or \answerNA{}.
    \item[] Justification: All details are provided in Section~\ref{sec:appendex_experiments}.

\item {\bf Experiment statistical significance}
    \item[] Question: Does the paper report error bars suitably and correctly defined or other appropriate information about the statistical significance of the experiments?
    \item[] Answer: \answerYes{} % Replace by \answerYes{}, \answerNo{}, or \answerNA{}.
    \item[] Justification: Significance details are provided in Section~\ref{sec:stdev}.

\item {\bf Experiments compute resources}
    \item[] Question: For each experiment, does the paper provide sufficient information on the computer resources (type of compute workers, memory, time of execution) needed to reproduce the experiments?
    \item[] Answer: \answerYes{} % Replace by \answerYes{}, \answerNo{}, or \answerNA{}.
    \item[] Justification: Compute resource details are provided in Section~\ref{sec:compute_resources}.
    
\item {\bf Code of ethics}
    \item[] Question: Does the research conducted in the paper conform, in every respect, with the NeurIPS Code of Ethics \url{https://neurips.cc/public/EthicsGuidelines}?
    \item[] Answer: \answerYes{} % Replace by \answerYes{}, \answerNo{}, or \answerNA{}.
    \item[] Justification: All work follows the code of ethics.

\item {\bf Broader impacts}
    \item[] Question: Does the paper discuss both potential positive societal impacts and negative societal impacts of the work performed?
    \item[] Answer: \answerYes{} % Replace by \answerYes{}, \answerNo{}, or \answerNA{}.
    \item[] Justification: This is discussed in Section~\ref{sec:limit}.
    
\item {\bf Safeguards}
    \item[] Question: Does the paper describe safeguards that have been put in place for responsible release of data or models that have a high risk for misuse (e.g., pre-trained language models, image generators, or scraped datasets)?
    \item[] Answer: \answerNA{} % Replace by \answerYes{}, \answerNo{}, or \answerNA{}.
    \item[] Justification: This is not applicable.

\item {\bf Licenses for existing assets}
    \item[] Question: Are the creators or original owners of assets (e.g., code, data, models), used in the paper, properly credited and are the license and terms of use explicitly mentioned and properly respected?
    \item[] Answer: \answerYes{} % Replace by \answerYes{}, \answerNo{}, or \answerNA{}.
    \item[] Justification: The codebases we use are properly referenced in Section~\ref{sec:appendex_experiments}.

\item {\bf New assets}
    \item[] Question: Are new assets introduced in the paper well documented and is the documentation provided alongside the assets?
    \item[] Answer: \answerNA{} % Replace by \answerYes{}, \answerNo{}, or \answerNA{}.
    \item[] Justification: There are no newly released assets.

\item {\bf Crowdsourcing and research with human subjects}
    \item[] Question: For crowdsourcing experiments and research with human subjects, does the paper include the full text of instructions given to participants and screenshots, if applicable, as well as details about compensation (if any)? 
    \item[] Answer: \answerNA{} % Replace by \answerYes{}, \answerNo{}, or \answerNA{}.
    \item[] Justification: There are no experiments of this nature.

\item {\bf Institutional review board (IRB) approvals or equivalent for research with human subjects}
    \item[] Question: Does the paper describe potential risks incurred by study participants, whether such risks were disclosed to the subjects, and whether Institutional Review Board (IRB) approvals (or an equivalent approval/review based on the requirements of your country or institution) were obtained?
    \item[] Answer: \answerNA{} % Replace by \answerYes{}, \answerNo{}, or \answerNA{}.
    \item[] Justification: This is not applicable.

\item {\bf Declaration of LLM usage}
    \item[] Question: Does the paper describe the usage of LLMs if it is an important, original, or non-standard component of the core methods in this research? Note that if the LLM is used only for writing, editing, or formatting purposes and does \emph{not} impact the core methodology, scientific rigor, or originality of the research, declaration is not required.
    %this research? 
    \item[] Answer: \answerNA{} % Replace by \answerYes{}, \answerNo{}, or \answerNA{}.
    \item[] Justification: LLMs were not used in any aspect of conducting this research.

\end{enumerate}

\end{document}